\documentclass[sigconf]{acmart}

\usepackage{microtype}
\usepackage{graphicx}
\usepackage{subcaption}
\usepackage{booktabs}
\usepackage{hyperref}

\usepackage{multicol}
\usepackage{multirow}
\usepackage{colortbl}
\usepackage{makecell}

\newcommand{\ie}{\emph{i.e.}}     % notation of `i.e.`
\newcommand{\eg}{\emph{e.g.}} 
\usepackage[capitalize,noabbrev]{cleveref}
\usepackage{comment}

\newcommand{\lyw}[1]{\textcolor{black}{#1}}
\newcommand{\lywre}[1]{\textcolor{black}{#1}}
\newcommand{\lywcr}[1]{\textcolor{black}{#1}}
\newcommand{\tao}[1]{\textcolor{black}{#1}}

\newcommand{\rynq}{\textcolor[rgb]{0.,0,0}}
\AtBeginDocument{%
  }

\renewcommand\footnotetextcopyrightpermission[1]{} % removes footnote with conference information in first column
\setcopyright{none}
\setcopyright{acmlicensed}
\copyrightyear{2026}
\acmYear{2026}
\setcopyright{cc}
\setcctype{by}
\acmDOI{10.1145/3767308.3836145}
\acmConference[MM '26]{Proceedings of the 34th ACM International Conference on Multimedia}{November 10--14, 2026}{Rio de Janeiro, Brazil}
\acmBooktitle{Proceedings of the 34th ACM International Conference on Multimedia (MM '26), November 10--14, 2026, Rio de Janeiro, Brazil}
\acmISBN{979-8-4007-2213-4/2026/11}

\begin{document}

%%
%% The "title" command has an optional parameter,
%% allowing the author to define a "short title" to be used in page headers.
\title{Glass Surface Detection Grounded in 3D Visual Geometry}

%%
%% The "author" command and its associated commands are used to define
%% the authors and their affiliations.
%% Of note is the shared affiliation of the first two authors, and the
%% "authornote" and "authornotemark" commands
%% used to denote shared contribution to the research.
\author{Yiwei Lu}
\authornote{Both authors contributed equally to this research.}
\email{6243112037@stu.jiangnan.edu.cn}
\orcid{0009-0004-9311-8645}
\affiliation{%
  \institution{Jiangnan University}
  \city{Wuxi}
  \country{China}
}

\author{Ke Xu}
\authornotemark[1]
\email{kkangwing@gmail.com}
\orcid{0000-0001-5855-3810}
\affiliation{%
  \institution{University of Science and Technology of China}
  \city{Hefei}
  \country{China}
}
\author{Tao Yan}
\authornote{Tao Yan is the corresponding author.}
\email{yantao.ustc@gmail.com}
\orcid{0000-0002-9162-8551}
\affiliation{%
  \institution{Jiangnan University}
  \city{Wuxi}
  \country{China}
}
\author{Xiaojun Chang}
\email{cxj273@gmail.com}
\orcid{0000-0002-7778-8807}
\affiliation{%
  \institution{University of Science and Technology of China}
  \city{Hefei}
  \country{China}
}

\author{Radu Timofte}
\email{radu.timofte@uni-wuerzburg.de}
\orcid{0000-0002-1478-0402}
\affiliation{%
  \institution{University of Wurzburg}
  \city{Bayern}
  \country{Germany}
}
\author{Rynson W. H. Lau}
\email{rynsonwhlau@gmail.com}
\orcid{0000-0002-8957-8129}
\affiliation{%
  \institution{City University of Hong Kong (Dongguan)}
  \city{Dongguan}
  \country{China}
}

\renewcommand{\shortauthors} {Yiwei Lu et al.}

%%
%% The abstract is a short summary of the work to be presented in the
%% article.
\begin{abstract}
Glass surface detection (GSD) is critical for scene understanding and reconstruction, and yet remains challenging due to the transparency and reflectivity of glass surfaces. 
Existing GSD methods typically rely on 2D appearance cues, which may fail in geometrically ambiguous scenes.
In this paper, we propose a paradigm shift: grounding GSD in 3D visual geometry to explicitly model the physical existence of glass surfaces.
Our method first distills rich 3D priors from the visual geometry grounded transformer (VGGT) and generates glass-aware 3D representations. It then exploits multi-tasking learning with a novel glass detection head, consisting of two core modules: a Frequency Self-Attention Module (FSAM) that identifies glass-specific spectral features for glass surface localization, and a Geometry Grounding Block (GeGB) that selectively grounds 2D features in 3D geometry for glass surface segmentation.
Extensive experiments demonstrate that our method achieves state-of-the-art performance across seven standard GSD benchmarks, generalizes well to video/multi-modal data, and substantially improves reconstruction in glass scenes.
\lywcr{Code is available in \url{https://github.com/YT3DVision/VGGT_GLASS}.}

\end{abstract} 

%%
%% The code below is generated by the tool at http://dl.acm.org/ccs.cfm.
%% Please copy and paste the code instead of the example below.
%%

%%
%% Keywords. The author(s) should pick words that accurately describe
%% the work being presented. Separate the keywords with commas.
%% A "teaser" image appears between the author and affiliation
%% information and the body of the document, and typically spans the
%% page.
% \begin{teaserfigure}
%  \includegraphics[width=\textwidth]{sampleteaser}
%   \caption{Seattle Mariners at Spring Training, 2010.}
%   \Description{Enjoying the baseball game from the third-base
%   seats. Ichiro Suzuki preparing to bat.}
%   \label{fig:teaser}
% \end{teaserfigure}

% \received{20 February 2007}
% \received[revised]{12 March 2009}
% \received[accepted]{5 June 2009}

%%
%% This command processes the author and affiliation and title
%% information and builds the first part of the formatted document.
\maketitle

\section{Introduction}
Glass is ubiquitous in our daily lives.
However, the transparency and specularity of glass create a unique form of semantic and geometric ambiguity for machine perception: the visual data of a glass surface contains both transmitted background scenes and foreground reflections.
Such inconsistency violates assumptions of many vision pipelines, \eg, depth/optical flow estimation~\cite{wen2025seeing,Poggi_2025_ICCV}, 3D reconstruction~\cite{ye20153d, reconstruction}, and autonomous navigation~\cite{autoRecons,weerakoon2024}.
Hence, accurate glass surface detection is essential for robust visual understanding and interactions in the real world.

%and is widely used in architectural materials such as windows and doors. However, its lack of distinctive visual characteristics poses challenges for many vision-based applications, including depth estimation, optical flow estimation, 3D reconstruction, and autonomous driving. This makes accurate glass surface detection essential.

%With the advancement of deep learning, numerous methods have been proposed to address this problem. 
Existing glass surface detection (GSD) methods typically learn appearance cues from the RGB domain (\eg, multiscale contextual features~\cite{mei2020GDNet,yu2022pgs, cheng2026}, reflections~\cite{lin2021GSDNet, Liu24videoglass}, ghosting effects~\cite{yan2025GhostingNet}, and boundaries~\cite{he2021EBLNet,fan2023RFENet}) or multi-modalities (combining RGB with, \eg, polarization~\cite{mei2022glass,qiao2023polarvideo}, thermal~\cite{huo2023glass}, and depth~\cite{lin2025rgbdglass}).
%near-infrared~\cite{yan2024nrglassnet}
%, and blurriness~\cite{Qi2024blur_glass}.
%Some multi-modal–based methods introduce additional modalities—such as polarization~\cite{mei2022glass, qiao2023polarvideo}, thermal infrared~\cite{huo2023glass}, depth~\cite{lin2025rgbdglass}, or near-infrared~\cite{yan2024nrglassnet}—to exploit the physical properties of glass surfaces and improve detection accuracy. 
%However, due to the limited size of existing datasets, these approaches often suffer from insufficient generalization ability. 
Most recently, \citet{li2025glasswizard} proposed a diffusion-based method to leverage the 2D image priors from large-scale foundation models for GSD.
%a unified glass surface detection network that achieved state-of-the-art performance across multiple datasets. This demonstrates that a single large model can be far more effective than training separate smaller models on individual datasets. 
Nonetheless, existing methods are inherently limited to 2D appearance-based reasoning: while effective in localizing glass regions, they tend to fail in segmenting glass surfaces accurately. As shown in Fig.~\ref{fig:instrction_compare} (2nd and 3rd columns), state-of-the-art methods typically produce false positives in surrounding non-glass regions that are visually similar to glass regions.
%glass surfaces do not have obvious reflections and the transmitted backgrounds are visually similar to surrounding non-glass regions.
%exhibit weak or ambiguous visual patterns, as shown in Fig.~\ref{fig:instrction_compare}.

\newcommand{\newsubwidth}{0.196}
\begin{figure}[t]
	\renewcommand{\tabcolsep}{0.8pt}
	\renewcommand\arraystretch{0.6}
        \centering
            \begin{tabular}{ccccc}
                % scene 1
                \includegraphics[width=\newsubwidth\linewidth]{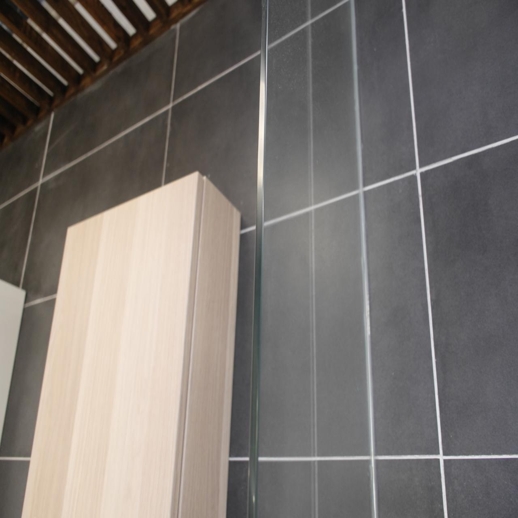}&
                \includegraphics[width=\newsubwidth\linewidth]{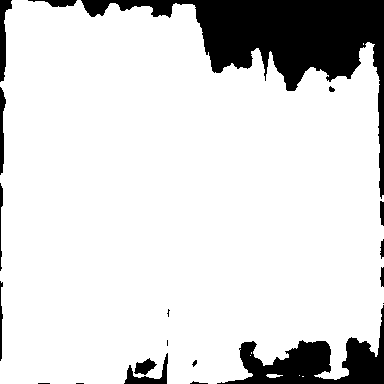}&
                \includegraphics[width=\newsubwidth\linewidth]{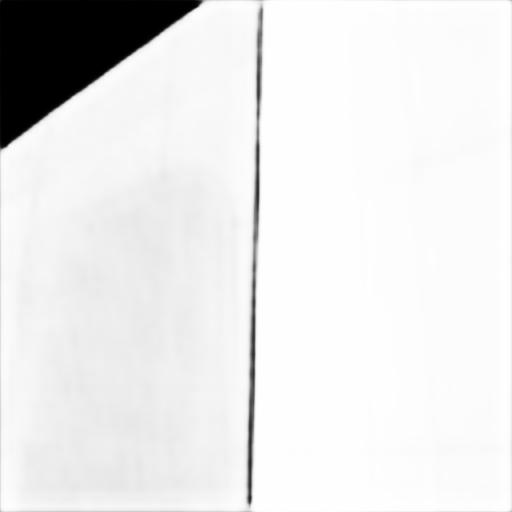}&
                \includegraphics[width=\newsubwidth\linewidth]{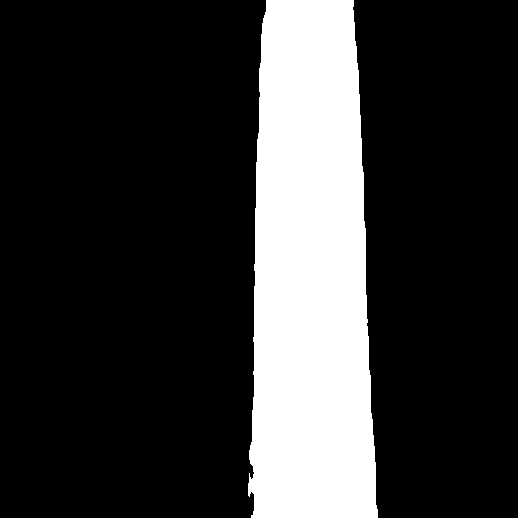}&
                \includegraphics[width=\newsubwidth\linewidth]{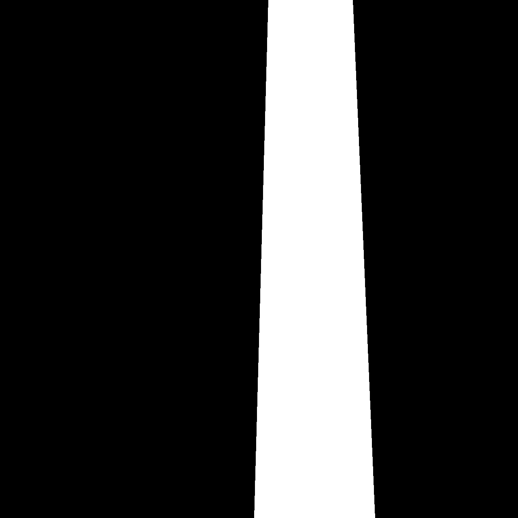}\\
  
                % scene 2
                \includegraphics[width=\newsubwidth\linewidth]{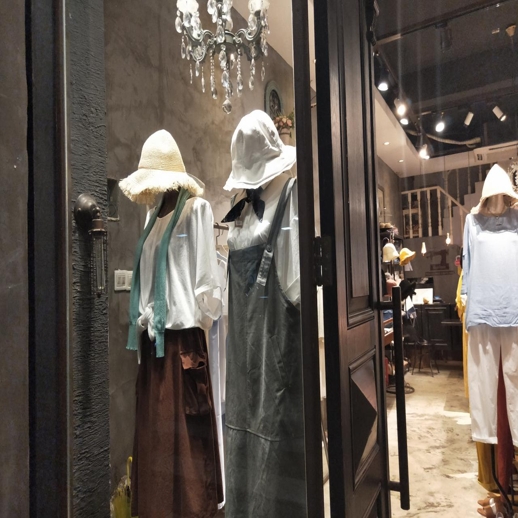}&
                \includegraphics[width=\newsubwidth\linewidth]{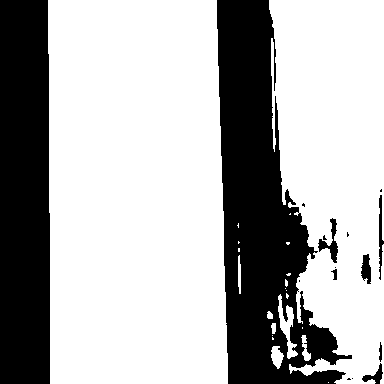}&
                \includegraphics[width=\newsubwidth\linewidth]{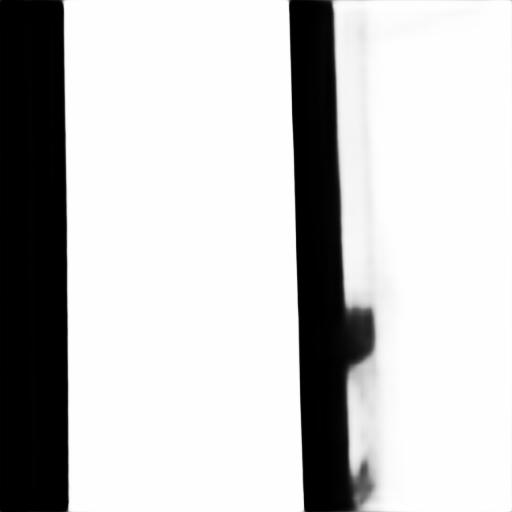}&
                \includegraphics[width=\newsubwidth\linewidth]{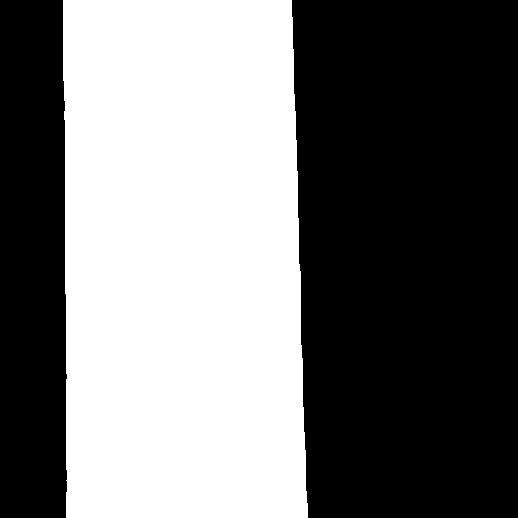}&
                \includegraphics[width=\newsubwidth\linewidth]{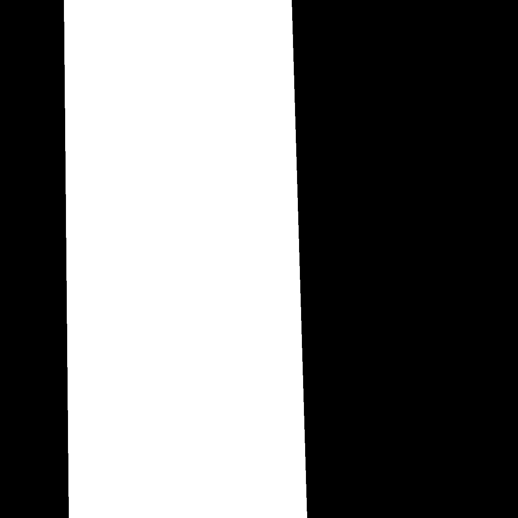}\\ 

                % annotation
                \small RGB&
                \small GhostingNet&
                \small GlassWizard&
                \small Ours&
                \small GT
            \end{tabular}
        % \vspace{-2mm}
        \caption{SOTA GSD methods typically rely on 2D appearance cues (\eg, by explicitly modeling the ghosting effects~\cite{yan2025GhostingNet}) or implicitly learning glass-specific embeddings from other modalities (\eg, text) to query large-scale pre-trained image diffusion priors~\cite{li2025glasswizard}. However, glass surfaces do not always exhibit strong ghosting effects~\cite{yan2025GhostingNet}. While the text-based image priors~\cite{li2025glasswizard} show promising glass surface localization performances, they are not sufficiently discriminative for pixel-wise segmentation. In this work, we propose to shift the paradigm by grounding GSD in large-scale pre-trained 3D visual geometry. We explicitly model the physical existence of glass surfaces, resulting in accurately glass segmentation.}
        \label{fig:instrction_compare}
        % \vspace{-3mm}
\end{figure}

We observe that humans resolve such visual ambiguity not by scrutinizing image pixels alone, but by constructing and reasoning within a 3D spatial model of the scene, as the intrinsic constraint theory~\cite{domini2023case} suggests: ``{\it human visual systems seek stable 3D interpretations from ambiguous 2D cues}''.
Glass surfaces, as physical objects, introduce geometric (\eg, depth and surface normals) and multi-view inconsistencies.
%hou2021pri3d, , yue2024fit3d
This motivates us to propose the paradigm shift, from modeling appearance-based cues to explicitly reasoning scene geometry, directly detecting and interpreting these systematic geometric inconsistencies as reliable signatures of glass surfaces by a 3D-grounded model.

%Fundamentally, humans reliably distinguish glass from non-glass regions by leveraging robust spatial perception and scene understanding—capabilities that current deep learning methods often lack. As noted by Lin et al.~\cite{lin2025rgbdglass}, the presence of glass surfaces often leads to missing regions in depth maps.
% , allowing depth information to reveal the potential location of glass. 
%Furthermore, Wen et al.~\cite{wen2025seeing} observed that representing scenes containing transparent objects with only a single-layer depth is insufficient, necessitating multi-layer depth information for accurate estimation. Motivated by these insights, we hypothesize that incorporating 3D geometric information can significantly enhance the robustness of glass surface detection.

We therefore turn to Visual Geometry Grounded Transformers (VGGT)~\cite{wang2025vggt}, a foundation model that encodes rich, large-scale 3D prior knowledge. Trained for multi-view reconstruction, VGGT can infer a comprehensive set of 3D attributes, including camera pose, depth, and point clouds, from single images with impressive generalization.
However, directly applying VGGT for GSD is non-trivial due to two core challenges: (1) VGGT inherently ``sees through'' glass, accurately modeling the occluded background \tao{while} failing to register the glass surface itself, and (2) a fundamental domain gap \tao{exists} between extracted 3D geometry features and 2D semantic features for pixel-wise glass surface segmentation.

%for 3D reconstruction that can predict a full set of 3D attributes, including camera parameters, depth maps, point maps, and 3D point trajectories, from an arbitrary number of input images. Despite being trained on large-scale video datasets, VGGT demonstrates strong generalization ability even with single-image input, largely due to the rich 3D prior knowledge embedded in the model. This allows our framework to rectify depth inaccuracies in glass regions, thereby providing more reliable geometric features for glass surface detection.

%Despite the impressive capabilities of VGGT, its direct application to the GSD task is hindered by three core challenges. First, VGGT inherently `sees through' glass, accurately modeling the background while failing to register the depth of the glass surface itself. Second, its architecture is optimized for multi-view consistency, introducing significant redundancy and excessive computational overhead when adapted for single-image inference. Third, there is a fundamental discrepancy between the 3D geometric features and the 2D semantic features.
% grounds in explicit 3D visual geometry
To address these challenges, we propose a novel framework that~\lywcr{detects glass surfaces based on a strong 3D vision foundation model.} Our approach has two steps.
First, we formulate a training pipeline that leverages VGGT to generate pseudo-ground-truth 3D scene \lywre{representations, and then refine these representations using a planar interpolation strategy to obtain more consistent geometric supervision.} 
% which \rynq{we rectify in glass regions (*** I am not too sure what you mean here. ***)} via a robust planar interpolation algorithm.
%
Second, we propose a glass detection head with two \lywre{complementary} components: the Frequency Self-Attention Module (FSAM) and the Geometry Grounding Block (GeGB). \lywre{FSAM captures ambiguous appearance patterns of transparent materials in the frequency domain, providing cues for localizing regions where standard spatial features are unreliable. Building upon this, GeGB incorporates rectified 3D geometric features to produce geometry-aware representations for accurate glass segmentation.}
% \rynq{FSAM learns to decouple glass-specific appearance features (characterized by high-frequency attenuation) to localize glass regions. (*** Is this related to 3D? Since the main idea of this work is to consider GSD in 3D, it is better for the claimed contributions to be related to 3D too. Otherwise, it can be difficult to claim this module as novel. ***)} GeGB then enriches the learned FSAM features with rectified 3D geometry features, producing local geometry-conditioned features for glass surface segmentation.
%
Our method is efficiently fine-tuned via Low-Rank Adaptation (LoRA), preserving the foundational 3D knowledge while specialized for GSD.
As shown in Fig.~\ref{fig:instrction_compare} (4th column), our method can accurately differentiate glass regions from ambiguous non-glass ones.
%To address these challenges, we propose VGGT-S, a novel baseline that explicitly integrates 3D visual geometric priors into the GSD task. Specifically, we utilize a 3D visual geometry encoder to generate comprehensive scene descriptors, which are then decoded into depth maps, point clouds, glass masks and edge masks. 
%To bridge the gap between 3D geometry features and glass features, we introduce a cascaded glass head consisting of two key modules. First, the Frequency Self-Attention Module (FSAM) leverages frequency-domain attention to decouple subtle glass features from dense 3D priors. Subsequently, the Multi-Modal Fusion Block (MMFB) facilitates an adaptive feature-level alignment between 3D geometry and 2D appearance. 
%This modular design is inherently hot-swappable, allowing our glass detection head to be seamlessly integrated with diverse backbones to enhance their geometric awareness in challenging environments.

In summary, the main contributions of this work include:
\begin{itemize}
\item We propose to ground the glass surface detection task in 3D visual geometry and present the first framework that effectively leverages large-scale 3D geometry priors for robust glass surface detection, establishing a new strong baseline.
%\item We introduce VGGT-S, a new and strong baseline for glass surface detection that, for the first time, leverages rich 3D priors from a large pre-trained visual geometric model. Furthermore, we develop a two-step training pipeline that first corrects the three-dimensional information of the glass area and then effectively exploits it for robust glass surface detection.
\item We propose a novel glass head with two core modules: FSAM, which learns discriminative glass features in the frequency domain, and GeGB, which adaptively grounds learned 2D appearance features in 3D geometry priors.
% a frequency-domain distinction module that enriches representations by explicitly learning the discriminative spectral differences between glass and non-glass regions, and a geometry-guided enhancement module that highlights glass surfaces by leveraging depth and point-cloud features as strong 3D cues.
\item We highlight the effectiveness and geometric awareness of our learned GSD representations through extensive experiments, demonstrating SOTA performances on \rynq{seven standard benchmarks,} strong generalization to multi-modal or video GSD data, and substantial improvements on glass scene reconstruction.
%(*** Are you sure that we have really tested on seven benchmarks? ***)Extensive experiments show that our proposed method outperforms the SOTA methods in most datasets and generalizes well to unseen scenes.
\end{itemize}

\begin{figure*}[t]
    \centering
    \includegraphics[width=1.7\columnwidth]{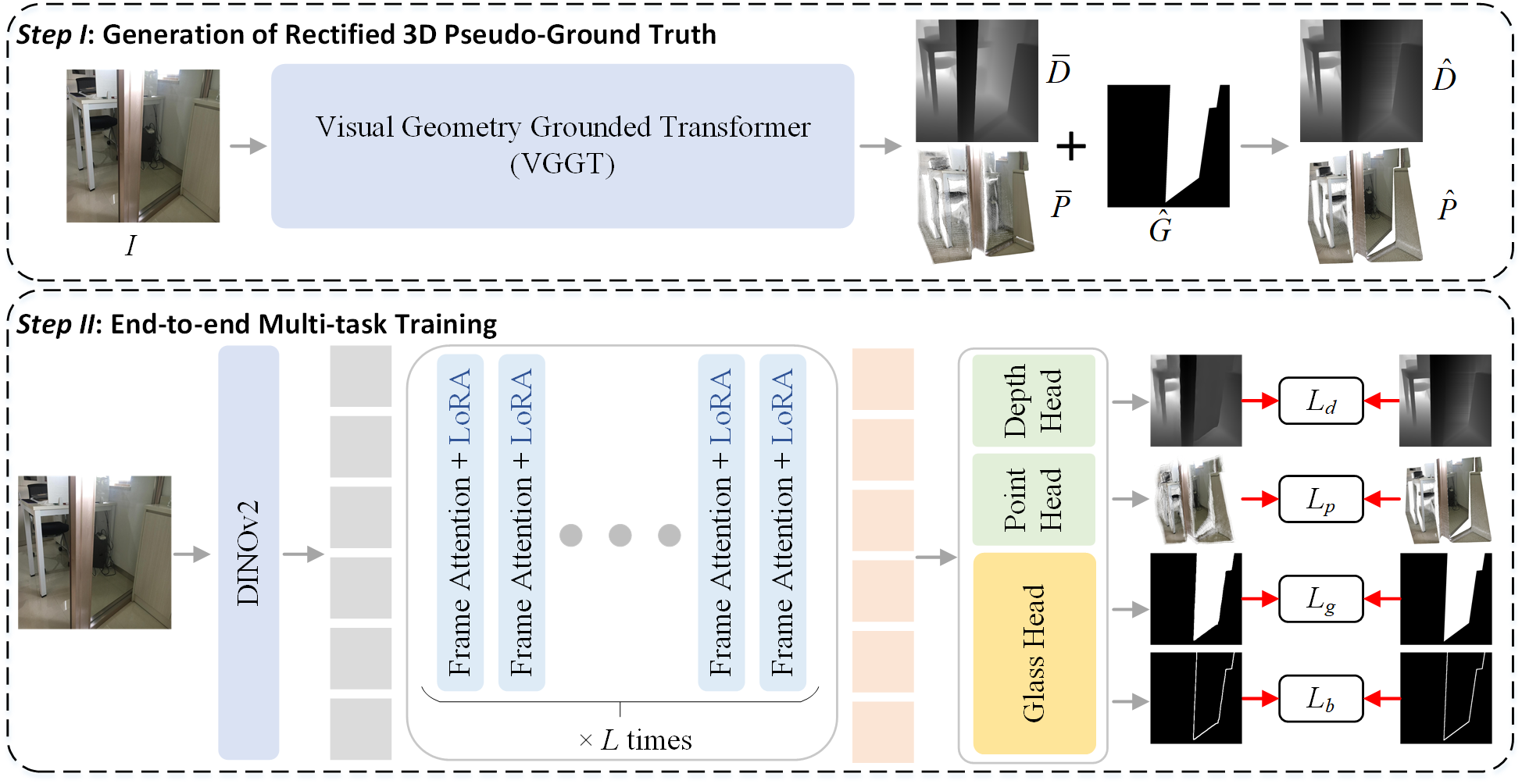}
    %\hfill
    %\vspace{-2mm}
    %\rynq{two-fold approach (*** This sounds a bit strange. What does this mean??? I suggest not to use "two-fold" here. Reviewers would not understand this. I understand "two-stream", but what is "two-fold approach"? ***)}
    \vspace{-2mm}
    \caption{Method Overview. We present a novel two-step approach that grounds glass surface detection in 3D visual geometry. In the first step, we generate rectified 3D pseudo-ground truth of point clouds and depths by leveraging VGGT and ground truth glass surface masks. In the second step, we formulate a multi-task learning objective with a novel glass head for geometry-aware GSD. Our method achieves state-of-the-art GSD performance, generalizes well to video/multi-modal data, and substantially improves reconstruction in glass scenes.}
    \label{fig:overview}
    \vspace{-3mm}
\end{figure*}

\section{Related Work}

\subsection{Glass Surface Detection (GSD)}
%Since glass is colorless, transparent, and lacks distinctive visual cues, it is difficult for vision systems to detect. Early works primarily addressed this problem from two directions: single-image cue mining and multimodal-assisted detection. 

Existing deep GSD methods can be broadly categorized into RGB-based and multi-modal approaches.
%Single-image methods improve detection accuracy by exploiting semantic correlations
RGB-based methods exploit various appearance cues, such as multiscale contextual features~\cite{mei2020GDNet,yu2022pgs, cheng2026}, boundary information~\cite{he2021EBLNet, fan2023RFENet}, reflections~\cite{lin2021GSDNet}, ghosting effects~\cite{yan2025GhostingNet}, and \lywre{temporal information}~\cite{Liu24videoglass, wang2025videoTOS, lu2026mvgd},
%, and blurriness~\cite{Qi2024blur_glass} within glass regions.
while multimodal methods incorporate additional modalities such as polarization~\cite{mei2022glass, qiao2023polarvideo}, thermal infrared~\cite{huo2023glass}, near-infrared~\cite{yan2024nrglassnet}, and depth~\cite{lin2025rgbdglass} to complement the RGB-based appearance cues.
%assist glass surface detection, achieving remarkable performance. 
Very recently, two GSD methods based on visual foundation models have been proposed.~\citet{hao2025gem} generate large amounts of glass-containing images using the Stable Diffusion model~\cite{rombach2022stable} to train the Segment Anything Model (SAM)~\cite{kirillov2023sam} for GSD, while~\citet{li2025glasswizard} propose the Stable Diffusion-based GSD model.

% Despite their success
Despite their success, these methods primarily relying on 2D appearance cues struggle in scenes where glass surfaces do not exhibit distinct visual patterns. In this paper, we show that our method achieves robust detection by grounding GSD in 3D visual geometry. 

%We note there is a concurrent work 
%The \tao{method mostly related to our method} is MonoGlass3D~\cite{zhang2025monoglass3d}, which acquires point cloud representations of glass scenes using LiDAR and annotates glass regions in 3D space to enable the model to learn geometric representations of glass. Our work differs significantly in two ways. First, our work is built on the visual foundation model VGGT, which provides strong generalization capabilities. Second, through knowledge distillation, our method enables the model to learn geometric information about the scene without relying on any additional data.

\subsection{Low Rank Adaptation (LoRA)}
The rapid scaling of visual foundation models has made parameter-efficient-fine-tuning (PEFT) essential for adapting large pre-trained models to downstream tasks. Among PEFT methods, LoRA~\cite{hu2022lora} emerges as a simple yet highly effective method. Instead of updating all model parameters, LoRA injects learnable low-rank matrices into the linear layers of a frozen backbone, which reduces the number of trainable parameters by orders of magnitude while preserving the pre-trained knowledge.
Subsequent works then improve LoRA for reducing memory usage~\cite{dettmers2023qlora} and numbers of trainable parameters~\cite{zhang2023lorafa,liu2024dora}, improving optimization dynamics~\cite{zhang2023adalora,wang2024loraga}, and enhancing LoRA's performance~\cite{huang2025hira,zhuang2025coto}. While these methods offer improvements, the vanilla LoRA formulation remains widely adopted due to its robustness and simplicity.

In our work, we adopt the vanilla LoRA to efficiently adapt the VGGT encoder to leverage 3D geometry priors for GSD. LoRA allows preserving VGGT's powerful 3D representations while specializing the model for GSD with minimal parameter overhead.

\section{Method}
% Section 3.1 introduces the overall framework, Section 3.2 details the pseudo ground truth, Section 3.3 describes the glass decoder along with the two proposed modules, and Section 3.4 presents the loss function.
%\subsection{Overview}
We aim to establish a new paradigm for glass surface detection by explicitly grounding this task in 3D geometry. The core insight is that glass surfaces, despite their visual ambiguity, introduce consistent geometric irregularities such as depth inconsistency and multi-view consistency, which can be reliably detected by models with strong 3D scene understanding priors.
To this end, we build our method upon the Visual Geometry Grounded Transformer (VGGT)~\cite{wang2025vggt} as a foundational source of large-scale 3D priors, and propose a novel method to adapt these priors for glass surfaces detection.

%Our \tao{method, named VGGT-S, fully exploits} 3D priors of visual geometric models \tao{for obtaining} more accurate results in glass surface detection. \tao{It leverages} 3D information in three ways: (1) our network is \tao{built upon} the visual geometry model VGGT pretrained on a large scale 3D reconstruction dataset, (2) we fine-tune the network's parameters \tao{to make it} detect glass surfaces without losing knowledge about 3D reconstruction, (3) we propose a novel feature fusion strategy. As illustrated in Fig.~\ref{fig:overview}, \tao{our method contains two main steps.} In step \uppercase\expandafter{\romannumeral1}, we use VGGT to generate pseudo ground-truth for the 3D information of the scene. In step \uppercase\expandafter{\romannumeral2}, we use joint training to train our network end-to-end.

\subsection{Pipeline}\label{sec:pipeline}

As shown in Fig.~\ref{fig:overview}, our method contains two steps: (1) generation of rectified 3D pseudo-ground truth and (2) end-to-end multi-tasking learning with a novel glass detection head. 
In the first step, we use VGGT to infer dense depth $\bar{D}\in\mathbb{R}^{H \times W}$ and point cloud $\bar{P}\in\mathbb{R}^{3 \times H \times W}$ from a single RGB image $I\in\mathbb{R}^{3 \times H \times W}$. Since VGGT ``sees through'' glass surfaces, we rectify the geometry in glass regions via a planar interpolation method based on the ground truth glass mask $\hat{G}\in\mathbb{R}^{H \times W}$ (Sec.~\ref{sec:depth_fix}).
In the second step, the input image $I\in\mathbb{R}^{3 \times H \times W}$ is encoded by the adapted VGGT backbone (consisting of $L$ frozen frame attention layers with learnable low-rank matrices in the linear layers for Low Rank Adaptation (LoRA)). 
The extracted features are then processed by three decoders for depth, point cloud, and glass (and edge) prediction. A novel glass head (Sec.~\ref{sec:glass_head}) is proposed to segment glass surfaces and delineate their boundaries based on 3D geometric and 2D appearance features.

\begin{figure}[t]
\renewcommand{\newsubwidth}{0.23}
\renewcommand{\tabcolsep}{0.6pt}
\renewcommand\arraystretch{0.6}
    \centering
    \small
    \begin{tabular}{cccc} 
        % 531 image
        \includegraphics[width=\newsubwidth\linewidth]{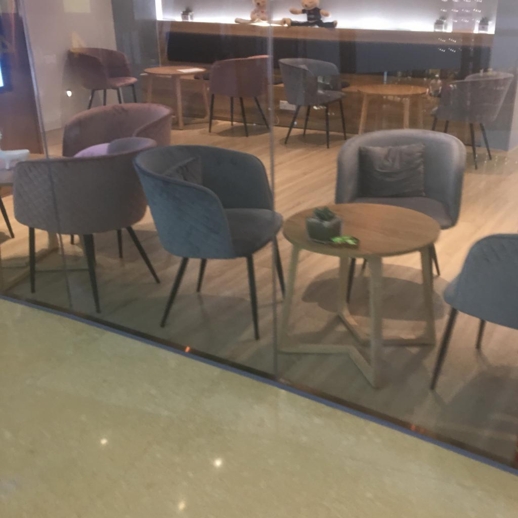}&
        \includegraphics[width=\newsubwidth\linewidth]{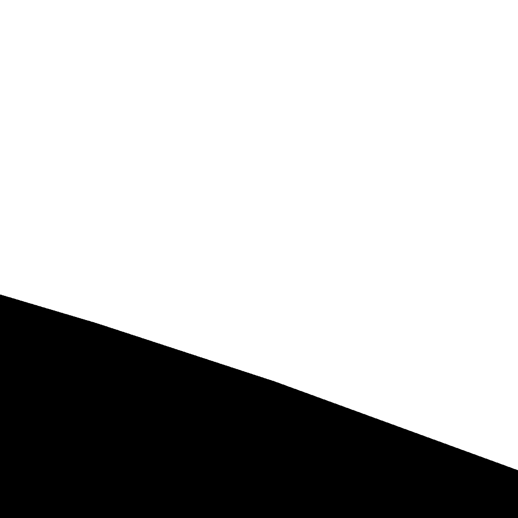}&
        \includegraphics[width=\newsubwidth\linewidth]{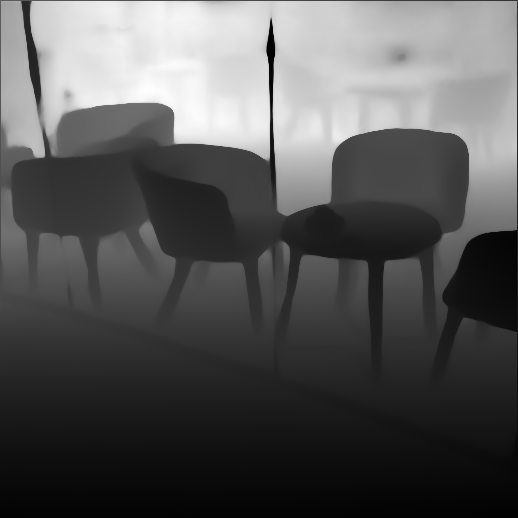}&
        \includegraphics[width=\newsubwidth\linewidth]{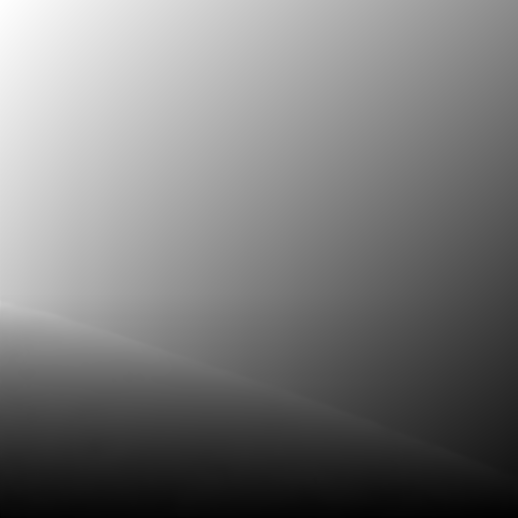}
        \\ 
        \includegraphics[width=\newsubwidth\linewidth]{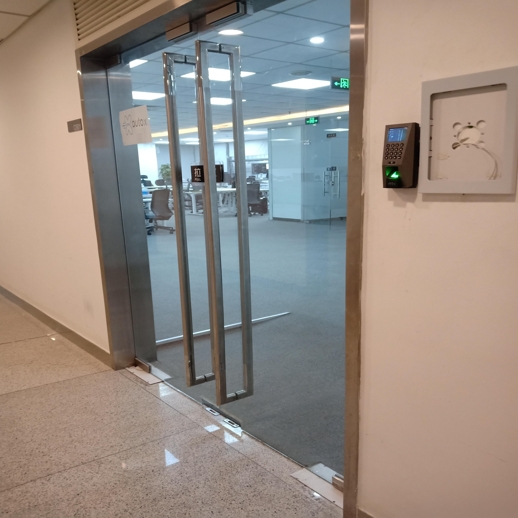}&
        \includegraphics[width=\newsubwidth\linewidth]{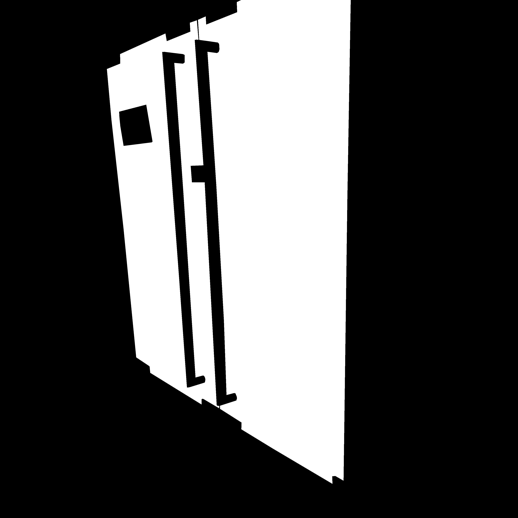}&
        \includegraphics[width=\newsubwidth\linewidth]{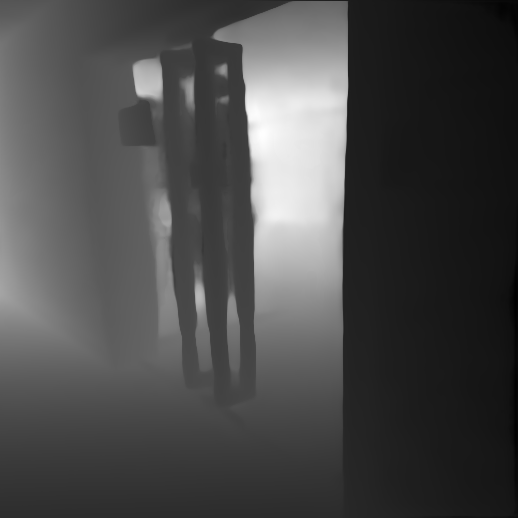}&
        \includegraphics[width=\newsubwidth\linewidth]{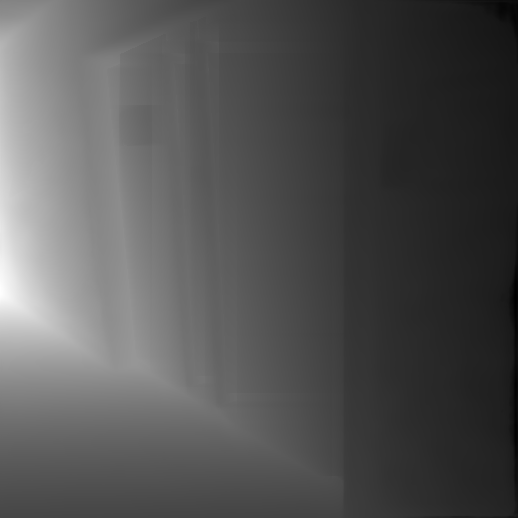}
        \\ 
        RGB&Glass&Raw Depth&Glass's Depth
    \end{tabular}
    \hfill
    \vspace{-2mm}
    \caption{Depth map correction for glass surfaces improves their physical plausibility and geometric coherence.}
    \label{fig:depth_samples}
    \vspace{-3mm}
    %\caption{Examples of corrected glass surfaces’ depth, which strengthen the physical existence of glass surfaces.}
\end{figure}

\subsection{Depth and Point Cloud Correction}\label{sec:depth_fix}

VGGT can provide strong geometry priors, but cannot reliably perceive glass surfaces, as it reconstructs the occluded background rather than the physical glass planes. This results in depths and point clouds that are semantically correct but geometrically inconsistent in the glass regions, thereby introducing misleading signals during training.
We address this problem via a geometry-aware correction method that explicitly injects a consistent 3D structure into the glass regions.
Our method is motivated by the geometric observation that in most real-world scenes, glass surfaces, despite potential mild curvature, are approximately planar. This allows us to formulate the correction as a planar interpolation problem anchored at the glass boundary, where depth values are reliable in non-glass regions. This ensures that the corrected 3D data reflects both scene geometry and the physical presence of glass, providing coherent multi-modal learning signals for the model.

Specifically, given the initial depth map $\bar{D}$ and point cloud $\bar{P}$ from VGGT, and the ground truth glass mask $\hat{G}$, our method computes the corrected depth $\hat{D}$ and points $\hat{P}$ via:

%Encouraged by previous work~\cite{liang2023monodepth}, we also design an efficient depth interpolation method. Our method is grounded in the observation that glass surfaces, despite potential local curvatures, can be effectively modeled as planar entities within most indoor and outdoor scenes. To facilitate large-scale and efficient data processing, our method obviates the need for manual annotation of reliable points in the depth map. \tao{Instead}, we leverage the depth values at the glass boundaries as reliable anchors, assuming these regions are accurately estimated. 
%As shown in Fig.~\ref{fig:depth_fix}, Depth maps are rectified \tao{by our method} in two steps.
%First, glass surfaces usually appear only partially within the image, and the depth values at the image boundaries—especially at the four corners—are inherently unreliable. For these points on the corner, we assume that their depth is spatially constrained by the surrounding non-glass regions (e.g., the ground or the frame in which it is located). 
%Therefore, the depth of these points is set as a weighted sum of the depth of the nearest non-glass points and their own depth to retain a certain perspective effect. This process can be formulated as follows:

{\it (1) Boundary-aware Depth Anchoring.}
As pixels at boundaries and corners often lack geometric context, to preserve the global scene layout,we blend the original depth of a corner pixel with that of its nearest non-glass pixel to retain perspective consistency, as:
\begin{equation}
\hat{D}(i,j)= \alpha*\bar{D}(\text{Nearest}(i,j))+(1-\alpha)*\bar{D}(i,j),
\end{equation}
where $i\in\{0, H-1\}$ and $j\in\{0, W-1\}$ denote corner pixel coordinates, and $\text{Nearest}(\cdot)$ returns the coordinates of the nearest non-glass pixel to $(i,j)$. $\alpha$ is set to $0.8$ to bias the correction toward geometrically reliable non-glass regions.

{\it (2) Planar Interpolation within Glass Regions.}
For an interior glass pixel $(i, j)$, we model the local glass region as a plane defined by nearest non-glass depths along the horizontal and vertical directions. Let the pixel's vertical bounds be $a,b$ and horizontal bounds $c,d$, where $\bar{D}(a, j), \bar{D}(b, j),\bar{D}(i, c), \bar{D}(i, d)$ are depths from nearest non-glass pixels along the respective directions. We compute the corrected depth $\hat{D}(i,j)$ as:
%Second, for the remaining points, we estimate the depth of each point using the depths of its four nearest non-glass neighbors along the horizontal and vertical directions. For example, the depth $\hat{D}(i, j),i\in[a,b],j\in[c, d]$ is obtained by averaging the results of two interpolation algorithms: one using two non-glass points ($\bar{D}(i, c), \bar{D}(i, d)$) in the horizontal direction and the other using two non-glass points ($\bar{D}(a, j), \bar{D}(b, j)$) in the vertical direction. The specific formula is as follows:
\begin{align}
\hat{D}_{v}(i,j)= \bar{D}(a, j)+\frac{i-a}{b-a}(\bar{D}(b, j)-\bar{D}(a, j)),\\
\hat{D}_{h}(i,j)= \bar{D}(i, c)+\frac{j-c}{d-c}(\bar{D}(i, d)-\bar{D}(i, c)),\\
\hat{D}(i,j) = 0.5*\hat{D}_{v}(i,j)+0.5*\hat{D}_{h}(i,j).
\end{align}
Examples of the corrected depths are shown in Fig.~\ref{fig:depth_samples}. The corrected depths $\hat{D}$ are combined with initial point clouds $\bar{P}$ to produce $\hat{P}$.

\begin{figure}[h]
    \centering
    \includegraphics[width=0.9\columnwidth]{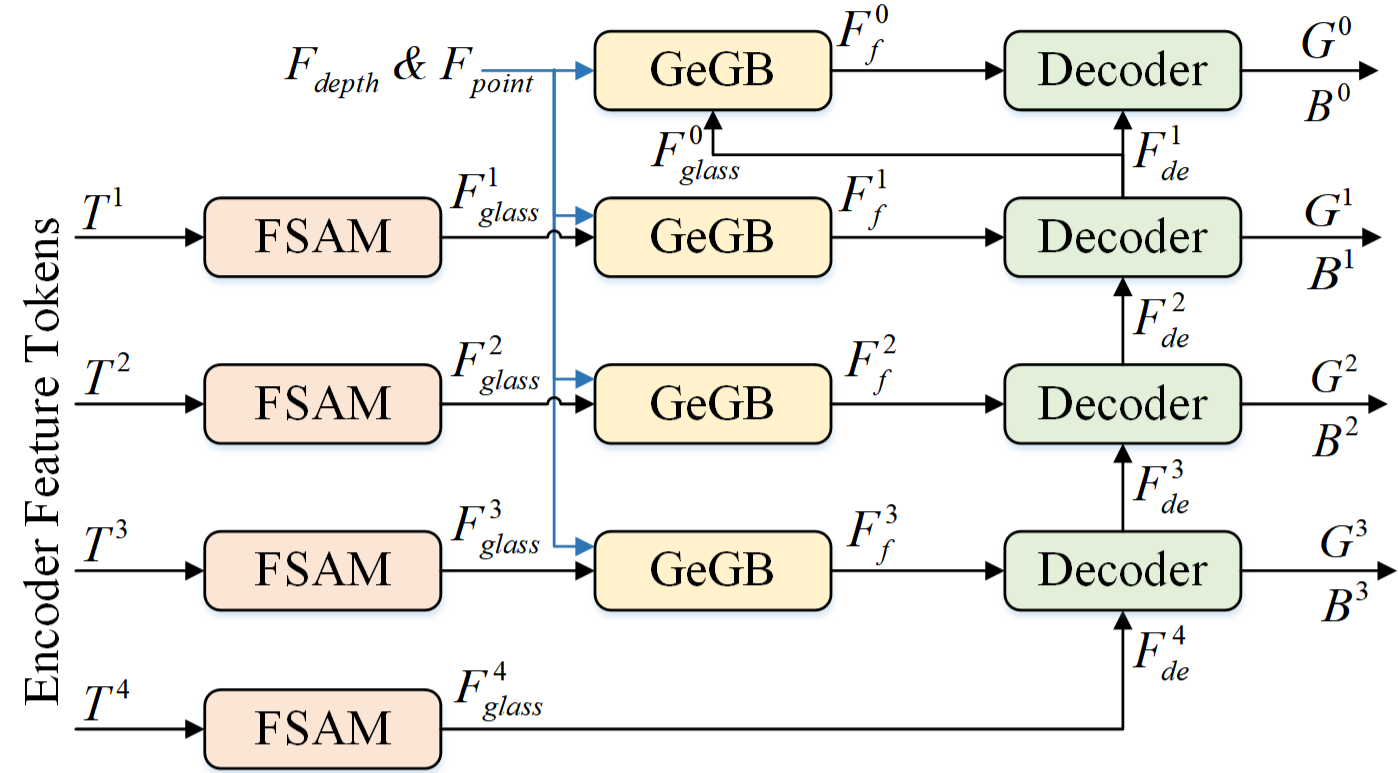}
    %\hfill
    \vspace{-2mm}
    \caption{Our Glass Head for precise glass surface segmentation and glass boundary delineation is built upon (1) the Frequency Self-Attention Module (FSAM) for capturing discriminative spectral representation of glass, and (2) the Geometry Grounding Block (GeGB) that reduces visual ambiguities via geometry-aware feature alignment and selective fusion.}
    % ~\kk{(**The multi-modal inputs of MMFB are not clear. Remove $F^{1,2,3,4}_{de}$ as they are not used.**)}
    \label{fig:glass_head}
    \vspace{-3mm}
\end{figure}

% \subsubsection{3D prediction.} 
%  These two heads share an identical network structure, distinguished only by their respective activation functions. Specifically, the depth detection head uses an inverse logarithmic transformation to obtain the depth map $D\in\mathbb{R}^{H \times W}$, while its confidence map $C_{d}\in\mathbb{R}^{ H \times W}$ is generated by exponential activation. Meanwhile, the point detection head uses exponential activation to produce both the point cloud $P\in\mathbb{R}^{3 \times H \times W}$ and its confidence map $C_{p}\in\mathbb{R}^{ H \times W}$. To ensure numerical stability during loss computation, a constant offset of $1$ is added to all confidence values.

\subsection{Glass Detection Head}\label{sec:glass_head}
We propose a novel glass detection head for precise glass surface segmentation and glass boundary delineation, based on two physically-grounded modules: \lywcr{a Frequency Self-Attention Module (FSAM) that captures glass-specific 2D visual features, and a Geometry Grounding Block (GeGB) that integrates 3D geometry features with 2D visual features for glass segmentation.}
% a Frequency Self-Attention Module (FSAM) that captures the discriminative spectral representation of glass, and a Geometry Grounding Block (GeGB) that performs geometry-aware feature alignment and gated fusion for reducing visual ambiguities.

%To predict the glass and edge masks, we design a specialized glass detection head consisting of two core modules. 

As shown in Fig.~\ref{fig:glass_head}, we first extract multi-scale token representations from the VGGT encoder at \{$4th$, $11th$, $17th$, $23rd$\} layers. We assign FSAM at each scale to learn glass-aware appearance features $\{F_{glass}^i\}_{i=1}^4$. \lywcr{$F_{glass}^4$ is directly used for the deepest decoding stage, whereas the other features} are then enriched with multi-modal point and depth features $(F_{point}, F_{depth})$ \tao{via} the GeGB, \tao{yielding the fused feature} $\{F_{f}^i\}_{i=1}^3$. \lywcr{Then}, the decoder takes $\{F_{f}^i\}_{i=1}^3$ as input and \lywcr{produces the decoded feature $\{F_{de}^i\}_{i=1}^3$ to predict} multi-scale glass surface masks $\{G^i\}_{i=1}^3$ and glass boundary maps $\{B^i\}_{i=1}^3$. \lywcr{Finally, $F_{de}^1$ are further refined by passing through the GeGB and decoder again to produce more fine-grained results.}

%As illustrated in Fig.~\ref{fig:glass_head}, the glass head extracts multi-level tokens from the $4th$, $11th$, $17th$, and $23rd$ encoder layers to capture a hierarchical representation of the scene. Since these features are shared across depth and point cloud estimation tasks, we introduce the Frequency Self-Attention Module (FSAM) to explicitly. These refined tokens are then reshaped into a multi-scale feature pyramid $\{F_{glass}^i\}_{i=1}^4 \in \mathbb{R}^{2^{i+5} \times \frac{H}{2^{i+1}} \times \frac{W}{2^{i+1}}}$ for dense prediction.
%Furthermore, motivated by the observation that 3D scene information can significantly improve glass surface detection, we propose a Multi-Modal Fusion Block (MMFB) to adaptively fuse the depth and point cloud features with the glass features. Finally, the fused features $\{F_{fusion}^i\}_{i=1}^4$ are fed into \tao{decoders} to predict the multi-scale glass masks $\{G^i\}_{i=1}^4$ and edge masks $\{B^i\}_{i=1}^4$ at different resolutions.

{\it Frequency Self-Attention Module (FSAM).}
Glass surfaces introduce subtle yet consistent appearance distortions due to light absorption, scattering, and refraction. Such glass-induced attenuation typically reduces local contrast and results in a characteristic high-frequency suppression that is largely irrelevant to the scene content.
We capture such appearance cues via FSAM in the frequency domain to preliminarily localize glass regions. 
\tao{Unlike standard spatial-domain self-attention, which may fail to distinguish glass features from complex backgrounds, our FSAM operates in the frequency domain to characterize the prominent spectral features inherent to glass surfaces.}
%While standard self-attention operates in the spatial domain, where glass features may easily be buried with complex background texture features, our FSAM computes attention in the frequency domain to model prominent spectral features of glass.
%The presence of glass can cause images to become blurry~\cite{Qi2024blur_glass} due to the physical absorption and scattering of light energy. This change is very subtle when there is no reflection on the glass or the reflection is weak. Unlike standard Gaussian blur, glass-induced blur as a reduction in local contrast: darker pixels tend to brighten while brighter pixels \tao{becoming} dim, leading to an overall attenuation of high-frequency components. \taoq{Based on this observation, we perform self-attention in the frequency domain to capture this change and propose the FSAM module.} 

\begin{figure}[t]
    \centering
    \includegraphics[width=0.9\columnwidth]{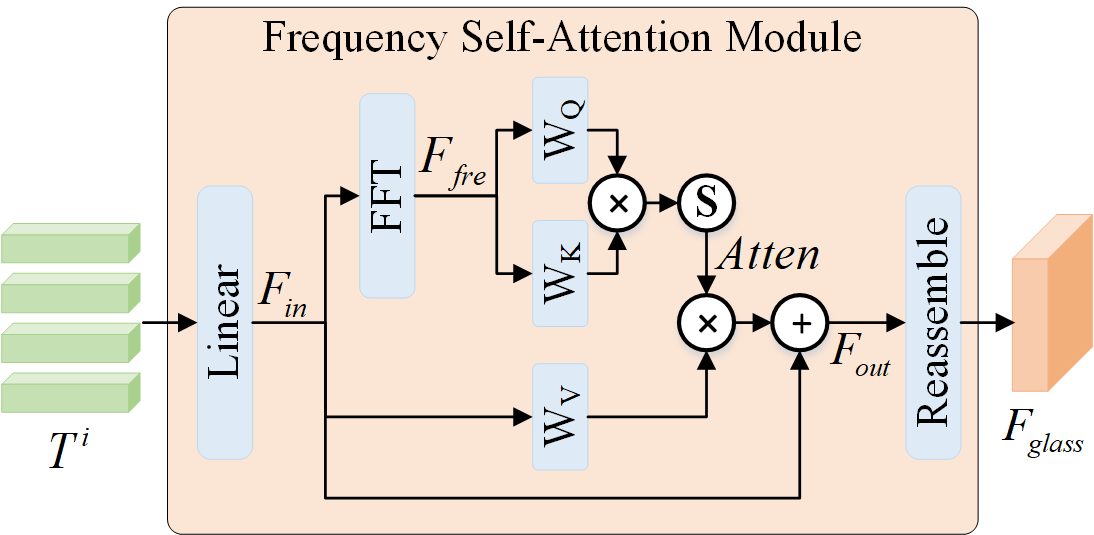}
    %\hfill
    \vspace{-2mm}
    \caption{Our Frequency Self-Attention Module (FSAM) operates in the frequency domain to capture prominent spectral features of glass surfaces for their preliminary localization.}
    \label{fig:fsam}
    \vspace{-3mm}
\end{figure}

As shown in Fig.~\ref{fig:fsam}, the extracted token representations at $i$-th scale, \tao{denoted as} {$\{T^i\}_{i=1}^4$}, are first projected to a lower-dimensional space via a linear layer to reduce computational cost. \tao{This process yields} $\{F^i_{in}\}_{i=1}^4 \in \mathbb{R}^{N \times C}$, where the output channel dimension is set to $C = 2^{i+5}$.
For each \lywcr{channel of} $F^i_{in}$\tao{,} we compute its amplitude spectrum via 1D Fast Fourier Transform (FFT) as \tao{follows}:
\begin{equation}\label{eq:fft}
    F_{fre} = \lvert \mathrm{FFT}(F_{in})\rvert.
\end{equation}
We then project $F_{fre}$ into query and key matrices via $W_Q(\cdot)$ and $W_K(\cdot)$, and compute the attention map:
\begin{equation}
    Atten = \mathrm{Softmax}(\frac{W_Q(F_{fre}) \times W_K(F_{fre})^{T}}{\sqrt{C}}).
\end{equation}
The frequency-domain attention map $Atten$ is then applied to the original spatial features $F_{in}$, followed by layer normalization and residual connection:
\begin{equation}\label{eq:attention}
    F_{out} = \mathrm{LN}(Atten \times W_V(F_{in}))+F^i_{in},
\end{equation}
where $W_V(\cdot)$ projects $F_{in}$ into value matrix. This formulation effectively circumvents the quantization error often caused by 1D Inverse Fast Fourier Transform (IFFT) and retains spatial information through residual connections.
Finally, the feature $F_{out}$ is re-shaped and processed via convolution and transposed convolution to produce the glass features $F_{glass}$ with aligned spatial and channel dimensions.

% FSAM is a simple yet effective design that exploits the spectral discriminability of glass surfaces by \lyw{performing} the attention computation \lyw{in} the frequency domain, effectively decoupling glass-specific features from complex background features.

%into a $C \times h \times w$ tensor, followed by convolution and transposed convolution to achieve the target spatial resolution, respectively.

%Inspired by the self-attention mechanism, we compute attention maps in the frequency domain and apply them back to the spatial domain.
%This cross-domain transformation enables the model to selectively enhance glass-specific features while strictly preserving the positional information of the feature vectors. The overall procedure can be formulated as follows:
%\begin{align}
%F_{fre} &= \mathrm{abs}(\mathrm{FFT}(F_{in})),\\
%Atten &= \mathrm{Softmax}(\frac{W_Q(F_{fre}) \times W_K(F_{fre})^{T}}{\sqrt{d}}),\\
%F_{out} &= \mathrm{LN}(Atten \times W_V(F_{in}))+F^i_{in},
%\end{align}
%where $d = C$ denotes the dimensionality of the attention embedding, $\mathrm{FFT}(\cdot)$ represents the Fast Fourier Transform, $\mathrm{abs}(\cdot)$ extracts the amplitude component by converting complex-valued outputs into real values, $\mathrm{Softmax}(\cdot)$ is applied to normalize the attention map, and $\mathrm{LN}(\cdot)$ denotes layer normalization. 
%Finally, the feature $F_{out}$ is re-projected into a $C \times h \times w$ tensor, followed by convolution and transposed convolution to achieve the target spatial resolution, respectively.

{\it Geometry Grounding Block (GeGB).}
While depth and point clouds both encode 3D geometry, they provide complementary information: depth maps offer dense pixel-aligned distance estimations, whereas point clouds capture local surface structures and provide volumetric context. Incorporating both modalities ensures that glass features are grounded in a comprehensive 3D scene representation.
%Due to the unique optical properties of glass, it exhibits significant visual variations in different images. Therefore, we decided to combine the geometric features of depth and point clouds with the semantic features of glass to improve the model's robustness to glass in different scenarios. Although both depth and point cloud features convey the geometric information of the scene, there are subtle differences between them. Our experiments demonstrate that utilizing both depth and point cloud features simultaneously yields the best results. 

\begin{figure}[t]
	\centering
    % \hspace{1pt} 
	\begin{subfigure}{0.53\linewidth}
		\centering
		\includegraphics[width=1.0\linewidth]{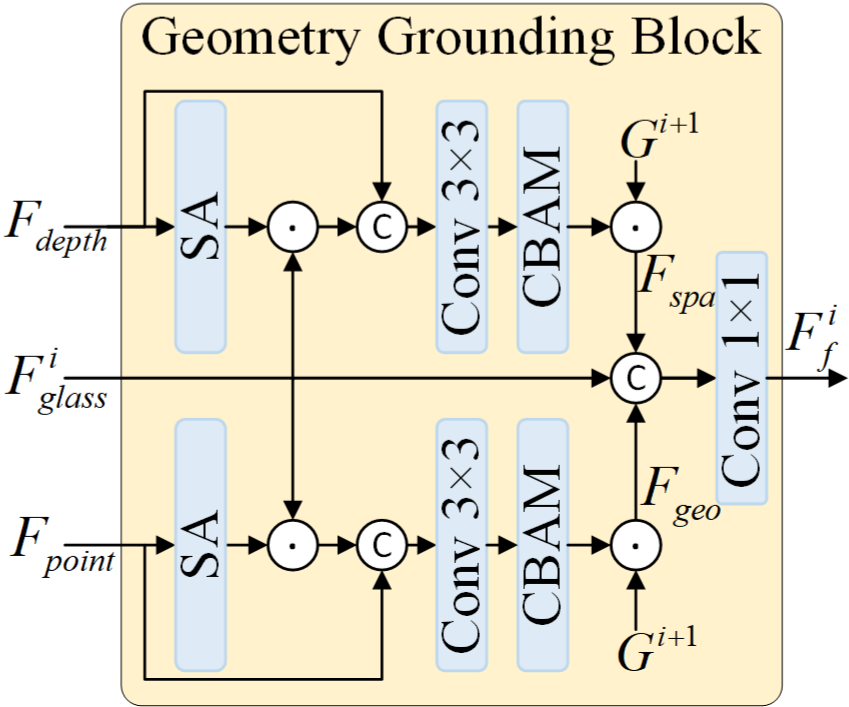}
		\caption{}
		\label{GeGB}
	\end{subfigure}
	\centering
    \hspace{-5.5pt} 
	\begin{subfigure}{0.47\linewidth}
		\centering
		\includegraphics[width=1.0\linewidth]{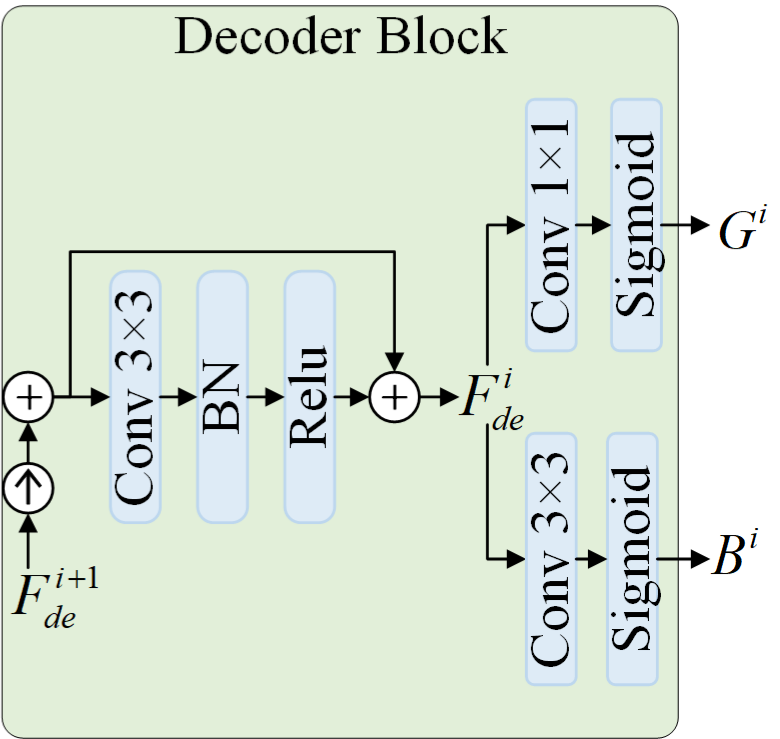}
		\caption{}
		\label{decoder}
	\end{subfigure}
	\caption{(a) Our Geometry Grounding Block (GeGB) incorporates depths and points as complementary geometric context for reasoning glass regions, and (b) our Decoder Block for segmenting glass regions and delineating their boundaries.}
	\label{fig:gegb_decoder}
    \vspace{-3mm} 
\end{figure}

To this end, we design a symmetrical gating-based geometry grounding block to incorporate both depth and point cloud features, as shown in Fig.~\ref{GeGB}.
%As shown in Fig~\ref{fig:mgfb}, we designed a vertically symmetrical structure to reflect the equal importance of depth and point cloud features. 
For each geometry modality $m\in\{depth, point\}$, we first compute a spatial attention map that highlights regions where geometric features are discriminative \tao{as follows}:
%use the Spatial Attention (SA) operation to obtain the spatial representation of the scene. This step can be expressed as:
\begin{equation}
\mathrm{SA}(F_{m})=\sigma(Conv_{7}([\text{Mean}(F_{m}), \text{Max}(F_{m})])),
\end{equation}
where $\mathrm{Mean}(\cdot)$ and $\mathrm{Max}(\cdot)$ compute the channel-wise average and maximum values, respectively, $[\cdot, \cdot]$ indicates feature concatenation, and $\sigma$ is sigmoid activation.
We then enrich each geometry modality with glass appearance features via a parallel enhancement path:
\begin{align}
F_{mm} = \mathrm{CBAM}(Conv_{3}([F_{m}, F^i_{glass} \odot \mathrm{SA}(F_{m})])),
%F_{geo}= \mathrm{CBAM}(Conv_{3}([F_{point}, F^i_{glass} \odot \mathrm{SA}(F_{point})])),
\end{align}
where $F_{mm}$ is either enhanced depth features $F_{spa}$ or point features $F_{geo}$, and CBAM~\cite{CBAM} is used to refine features via channel and spatial attention. 
%the glass features $F^i_{glass}$ are enriched with $F_{depth}$ to generate the spatial-enhanced feature $F_{spa}$. Similarly, $F_{point}$  is integrated to generate the geometric-enhanced feature $F_{geo}$.
We leverage the coarse glass surface prediction $G^{i+1}$ from the previous decoder layer \tao{to} guide the multi-modal fusion of $F^i_{glass}$, $F_{spa}$, and $F_{geo}$, as \tao{follows}:
%Finally, we used a $1 \times 1$ convolution layer to generate the fused feature $F^i_{fusion}$. This process can be formulated as:
\begin{align}
%F_{spa} = \mathrm{CBAM}(c^{3 \times 3}([F_{depth}, F^i_{glass} \odot \mathrm{SA}(F_{depth})])),\\
%F_{geo}= \mathrm{CBAM}(c^{3 \times 3}([F_{point}, F^i_{glass} \odot \mathrm{SA}(F_{point})])),\\
F^i_{f} = Conv_{1}[F^i_{glass},G^{i+1} \odot F_{spa},G^{i+1} \odot F_{geo}],
\end{align}
where $\odot$ is element-wise multiplication.

% \begin{figure}[t]
%     \centering
%     \includegraphics[width=1.0\columnwidth]{./fig/vggt_s/MMFB.png}
%     %\hfill
%     %\vspace{-2mm}
%     \caption{Left: our Geometry Grounding Block (GeGB). Right: our Decoder block.}
%     \label{fig:mgfb}
%     \vspace{-3mm}
% \end{figure}

%The attention map is then multiplied by the glass features to align the representations across modalities. Subsequently, the aligned features are fused using a $3 \times 3$ convolution layer, followed by a CBAM~\cite{CBAM} block to further enhance the feature representations. To suppress the influence of non-glass regions, we leverage the glass mask $G^{i+1}$ predicted by the previous decoder layer to constrain the fused features. In addition, the glass features $F^i_{glass}$ are enriched with $F_{depth}$ to generate the spatial-enhanced feature $F_{spa}$. Similarly, $F_{point}$  is integrated to generate the geometric-enhanced feature $F_{geo}$. These two complementary representations ensure that the model captures both the surface localization and the volumetric structure of the glass. Finally, we used a $1 \times 1$ convolution layer to generate the fused feature $F^i_{fusion}$. This process can be formulated as:
%\begin{align}
%F_{spa} = \mathrm{CBAM}(c^{3 \times 3}([F_{depth}, F^i_{glass} \odot \mathrm{SA}(F_{depth})])),\\
%F_{geo}= \mathrm{CBAM}(c^{3 \times 3}([F_{point}, F^i_{glass} \odot \mathrm{SA}(F_{point})])),\\
%F^i_{fusion} = c^{1 \times 1}[F^i_{glass},G^{i+1} \odot F_{spa},G^{i+1} \odot F_{geo}],
%\end{align}
%where $\odot$ means element-wise multiplication.

{\it Decoder Block.}
As shown in Fig.~\ref{decoder}, the decoder layer at each scale processes the $F^i_{f}$ and the $F_{de}^{i+1}$ from the previous layer via a standard residual block to produce $F_{de}^i$.
%progressively refines the features $\{F_{de}^i\}_{i=1}^4$. Specifically, at each stage, the decoder features $F_{de}^{i+1}$ are first up-sampled to match the resolution of the corresponding fused features $F_{fusion}^i$, followed by an element-wise summation for feature integration. Subsequently, a residual convolution block is employed to further refine the integrated features. 
We assign a $1 \times 1$ convolution layer to generate the glass mask and a $3 \times 3$ convolution layer to delineate glass boundaries.
% \begin{figure*}[!htp] 
%     \centering
%     \includegraphics[width=1.78\columnwidth]{./fig/vggt_s/frequency.png}
%     %\hfill
%     %\vspace{-2mm}
%     \caption{The architecture of our frequency module.}
%     \label{fig:frequency_moudle}
%     \vspace{-3mm}
% \end{figure*}

% \begin{figure*}[!htp]
%     \centering
%     \includegraphics[width=1.78\columnwidth]{./fig/vggt_s/decoder.png}
%     %\hfill
%     %\vspace{-2mm}
%     \caption{The architecture of our Decoder block.}
%     \label{fig:decoder_block}
%     \vspace{-3mm}
% \end{figure*}

\subsection{Multi-Modal Loss Function}\label{sec:loss_function}
We use a hybrid multi-task loss to train our network:
\begin{equation}\label{eq:loss}
\mathcal{L}=\mathcal{L}_{g}+\mathcal{L}_{b}+\lambda\mathcal{L}_{d}+\lambda\mathcal{L}_{p},
\end{equation}
where $\lambda$ is a balancing hyper-parameter empirically set to $0.1$. For geometry prediction, we adopt the same depth loss $\mathcal{L}_{d}$ and point cloud loss $\mathcal{L}_{p}$ as defined in VGGT~\cite{wang2025vggt}:
\begin{align}
\mathcal{L}_{d} = \| C_d \odot (D - \hat{D}  ) \|+  \|  C_d \odot  (\nabla D - \nabla\hat{D}  ) \|-\log_{}{C_d},\\
\mathcal{L}_{p} = \| C_p \odot  (P - \hat{P}  )   \|+  \|  C_p \odot  ( \nabla P - \nabla\hat{P} )   \|- \log_{}{C_p},
\end{align}
where $C_d$ and $C_p$ are the predicted confidence maps for depths and points, respectively. $D$ and $P$ are the predicted depth maps and point clouds, while $\hat{D}$ and $\hat{P}$ are our rectified pseudo ground truth depth and point. $\nabla$ computes gradients and $\odot$ is element-wise product. 
%Since the depth and point cloud detection heads are initialized with pre-trained parameters and their loss magnitudes are larger than those of the glass and edge detection heads, we introduce a scaling factor $\lambda = 0.1$ to down-weight their losses, encouraging the network to focus more on the glass surface detection task.

For glass surface predictions, we employ BCE loss~\cite{de2005CE} and IoU loss~\cite{qin2019BASNet} at the pixel- and region-level, respectively. Given the glass prediction $\{G^i\}_{i=0}^3$ and its corresponding ground truth mask $\hat{G}$, $\mathcal{L}_{glass}$ is defined as:
\begin{align}
\mathcal{L}_{glass}(G,\hat{G})=\mathcal{L}_{bce}(G,\hat{G})+\mathcal{L}_{iou}(G,\hat{G}),\\
\mathcal{L}_{g}=\sum_{i=1}^4\frac{\mathcal{L}^{'}_{glass}(G^i,\hat{G})}{2^{i-1}}+\mathcal{L}^{'}_{glass}(G^c,\hat{G}),
\end{align}
where $G^c$ is the glass surface mask derived from the point cloud and depth features (\ie, concatenation of $F_{depth}$ and $F_{point}$ followed by $1 \times 1$ convolution and Sigmoid activation), which facilitates the training via granular supervisions on geometry features.%\lyw{We discuss the impact of $G^c$ in the appendix.}
%, \lyw{which is obtained by concatenating the point cloud and depth features, followed by a $1 \times 1$ convolution and Sigmoid normalization. Compared to the direct output from $F^{4}_{glass}$, this mask maintains a superior spatial resolution of $518 \times 518$, thereby stabilizing the training process by providing more granular supervisory signals.}

For glass boundary predictions, $\mathcal{L}_{b}$ is defined in a similar way to $\mathcal{L}_{g}$, except that we replace the IoU loss with the Dice loss~\cite{milletari2016}.
\begin{table*}[t]
\centering
\setlength{\tabcolsep}{2.5pt}
% \small
\caption{Quantitative comparison between our method and existing GSD methods on standard single-image datasets. Foundation-model baselines are also included for reference. Best results are marked in {\bf bold}.}
% Quantitative comparison between our method and six SOTA methods on standard single-image GSD datasets. All methods are trained under the same setting
\begin{tabular}{lc ccccc c ccccc c ccccc}
\toprule
\multirow{2.45}{*}{Methods} &\multirow{2.45}{*}{Venue} &\multicolumn{5}{c}{GDD~\cite{mei2020GDNet}}&&\multicolumn{5}{c}{GSD~\cite{lin2021GSDNet}} &&\multicolumn{5}{c}{HSO~\cite{yu2022pgs}} \\
\cmidrule{3-7}  \cmidrule{9-13}  \cmidrule{15-19}
&&IoU$\uparrow$ &$F_\beta$$\uparrow$ &MAE$\downarrow$ &BER$\downarrow$ &ACC$\uparrow$ 
&&IoU$\uparrow$ &$F_\beta$$\uparrow$ &MAE$\downarrow$ &BER$\downarrow$ &ACC$\uparrow$ 
&&IoU$\uparrow$ &$F_\beta$$\uparrow$ &MAE$\downarrow$ &BER$\downarrow$ &ACC$\uparrow$\\
\midrule
SAM2 & -&
82.70&0.891&0.090&8.44&0.920 && 
73.00&0.797&0.096&9.24&0.917 && 
71.18&0.801&0.134&12.04&0.898\\
VGGT & CVPR'25&
85.26&0.914&0.072&6.68&0.948 && 
78.63&0.868&0.065&7.02&0.930 && 
77.62&0.854&0.094&8.35&0.944\\

SAM3 & ICLR'26&
92.88&0.966&0.034&3.04&0.974 && 
89.64&0.943&0.033&3.02&0.988 && 
87.24&0.931&0.054&4.66&0.965\\
\midrule
GDNet & CVPR'20& 
88.71&0.934&0.059&5.49&0.961 && 
83.00&0.893&0.055&5.68&0.952 && 
79.25&0.871&0.093&8.65&0.931 \\
GSDNet &CVPR'21& 
88.52&0.935&0.057&5.29&0.951 && 
82.96&0.895&0.052&6.00&0.938 && 
80.17&0.880&0.085&8.07&0.928\\
EBLNet & ICCV'21& 
88.02&0.932&0.059&5.51&0.950 && 
82.25&0.894&0.056&5.88&0.936 && 
80.53&0.887&0.082&7.79&0.927\\
RFENet & IJCAI'23& 
88.04&0.920&0.061&6.02&0.975 && 
81.26&0.869&0.049&6.17&0.961 && 
78.49&0.852&0.095&8.78&0.950\\
GhostingNet & TPAMI'25&
91.19&0.954&0.044&3.98&0.964 && 
86.69&0.923&0.042&4.19&0.972 && 
82.97&0.903&0.073&6.79&0.942\\
GlassWizard & ICCV'25&
93.30&0.969&0.039&3.62&0.967 && 
90.40&0.952&0.036&4.52&0.969 && 
87.90&0.941&0.055&5.44&0.952\\

% GlassWizard\cite{li2025glasswizard} & ICCV'25&93.89&0.969&0.039&3.62&0.962 && 91.76&0.952&0.036&4.02&0.952\\
%\midrule
%TOD & TransLab\cite{xie2020Translab} & ECCV'20&81.97&0.899&0.095&8.98&0.920 && 74.23&0.837&0.089&10.40&0.886\\
%TOD & Trans2Seg\cite{xie2021Transeg} & IJCAI'21&82.39&0.891&0.094&9.03&0.923 && 76.97&0.848&0.088&9.64&0.902\\
%\midrule
% VGGT~\cite{wang2025vggt}-LoRA & - &93.20& 0.964&0.031&3.20&0.965 && 89.60&0.939&{0.031}&{4.28}&{0.953}\\
% VGGT-S(baseline) & - &94.11&0.968&0.030&3.28&0.969 && 90.36&0.941&0.031&4.21&0.956\\
% VGGT-S(fix-depth) & - &94.54&0.971&0.029&3.01&0.967 && 90.86&0.945&0.031&4.34&0.943\\
% VGGT-S(ours) & - &\bf{95.58}&\bf{0.977}&\bf{0.025}&\bf{2.55}&\bf{0.971} && \bf{93.29}&\bf{0.957}&\bf{0.024}&\bf{3.21}&\bf{0.961}\\
\rowcolor[gray]{.8}{\bf Ours} & - &\bf{95.56}&\bf{0.981}&\bf{0.021}&\bf{1.88}&\bf{0.988} && \bf{91.91}&\bf{0.962}&\bf{0.022}&\bf{2.24}&\bf{0.989} && \bf{92.04}&\bf{0.963}&\bf{0.031}&\bf{2.73}&\bf{0.984}\\
% GSD & VGGT-S(ours) & - &{94.38}&{0.968}&\bf{0.029}&{3.05}&\bf{0.972} && {90.76}&{0.946}&0.034&4.63&0.938\\

% GSD & VGGT-G(New) & - &\bf{94.89}& \bf{0.973}&\bf{0.029}&\bf{2.87}&0.970 && 92.35/\bf{93.15}&0.952/\bf{0.956}&0.027/\bf{0.025}&3.85/\bf{3.45}&0.954/\bf{0.961}\\
\bottomrule
\end{tabular}
\label{tab:result}
\vspace{-2mm}
\end{table*}

\begin{table}[t]
\centering
\setlength{\tabcolsep}{3pt}
% \small
\caption{Quantitative comparison on the Trans10k-stuff. Best results are marked in {\bf bold}.}
\begin{tabular}{lc cccccc c}
\toprule
\multirow{2.45}{*}{Methods} &\multirow{2.45}{*}{Venue} &\multicolumn{5}{c}{Trans10k-stuff~\cite{xie2020Translab}} \\
\cmidrule{3-7} 
&&IoU$\uparrow$ &$F_\beta$$\uparrow$ &MAE$\downarrow$ &BER$\downarrow$ &ACC$\uparrow$\\
\midrule
SAM2 & -&
81.16&0.884&0.075&6.59&0.956\\
VGGT & CVPR'25&
84.86&0.912&0.060&5.39&0.963\\
SAM3 & ICLR'26&
91.07&0.956&0.033&2.82&0.984\\
\midrule
GDNet & CVPR'20& 
88.30&0.937&0.047&4.08&0.976\\
GSDNet &CVPR'21& 
88.62&0.940&0.043&3.87&0.974\\
EBLNet& ICCV'21& 
88.35&0.939&0.046&4.04&0.971\\
RFENet & IJCAI'23& 
86.51&0.912&0.053&4.84&0.992\\
GhostingNet & TPAMI'25&
89.72&0.948&0.038&3.40&0.979\\
GlassWizard & ICCV'25&
92.80&0.965&0.030&3.04&0.979\\
\rowcolor[gray]{.8}{\bf Ours} & - &
\bf{93.85}&\bf{0.974}&\bf{0.022}&\bf{1.76}&\bf{0.994}\\
\bottomrule
\end{tabular}
\label{tab:trans10k}
\vspace{-2mm}
\end{table}

\section{Experiments}

\subsection{Experimental Setups}
{\bf Implementation Details.}
Our framework is implemented with PyTorch on a single NVIDIA RTX 4090 GPU (24GB). We follow VGGT to use $L=24$ frame attention layers for feature extraction, where 
\lyw{Low Rank Adaptation (LoRA) is applied to each \tao{frame attention layer} with two low-rank matrices, $A \in \mathbb{R}^{d \times r}$ and $B \in \mathbb{R}^{r \times k}$ ($r \ll \min(d, k)$).}
%During the forward pass, the input is processed by both the original frozen weights and the product of these two small matrices ($A \times B$); the outputs are then summed.
%Specifically, we insert two matrices with a
We set the rank $r$ to $16$.
%projections to implement Low Rank Adaptation (LoRA).
% , where we integrate LoRA layers into the QKV projection matrices with a rank of $r=16$.
Images are resized to $518 \times 518$ for training and evaluation.
We train our model on a Union training set, which combines the training sets from GDD~\cite{mei2020GDNet}, GSD~\cite{lin2021GSDNet}, Trans10K-Stuff~\cite{xie2020Translab}, and HSO~\cite{yu2022pgs}. We employ the Adam optimizer with a learning rate of $1 \times 10^{-4}$ and a batch size of $4$. The training process spans $80$ epochs. 
% We will release codes for reproducibility.
%and notably, no data augmentation is utilized, further demonstrating the inherent representative power of our geometric priors.
% To train our model, we combine the training sets from GDD~\cite{mei2020GDNet}, GSD~\cite{lin2021GSDNet}, Trans10K-Stuff~\cite{xie2020Translab}, and HSO~\cite{yu2022pgs}. The merged dataset is termed as ``union glass''. All images are resized to $518\times518$ for both training and evaluation. Notably, no data augmentation is applied during training. Our model is implemented in PyTorch and trained on a single NVIDIA RTX 4090 GPU (24GB) with a batch size of $4$. The learning rate is set to $1\times10^{-4}$ with the Adam optimizer, and the model is trained for $80$ epochs. 

{\bf Evaluation Methods, Datasets, and Metrics.} 
We compare to nine existing GSD methods, including six single-image-based ones (GDNet~\cite{mei2020GDNet}, GSDNet~\cite{lin2021GSDNet}, EBLNet~\cite{he2021EBLNet}, RFENet~\cite{fan2023RFENet}, GhostingNet~\cite{yan2025GhostingNet} and GlassWizard~\cite{li2025glasswizard}), two multi-modal methods (RGB-Thermal~\cite{huo2023glass} and RGB-Depth~\cite{lin2025rgbdglass}), two video GSD methods (VGSDNet~\cite{Liu24videoglass} and MVGDNet~\cite{lu2026mvgd}). We include three foundation models (SAM2~\cite{ravi2024sam2}, VGGT~\cite{wang2025vggt}, and SAM3~\cite{carion2025sam3}) as baselines. \lywcr{For VGGT, we used the same segmentation decoder as ours.}
% \lywcr{SAM2 used 10 randomly sampled GT-mask points as prompts, while SAM3 adopted the text prompt “glass surface”. For VGGT, we used the pretrained encoder with the same segmentation decoder as ours.}

We evaluate on seven standard benchmarks, including four single-image-based datasets (GDD~\cite{mei2020GDNet}, GSD~\cite{lin2021GSDNet}, HSO~\cite{yu2022pgs} and Trans10k-stuff~\cite{xie2020Translab}), two multi-modal GSD dataset (RGB-Thermal-based~\cite{huo2023glass} and RGB-Depth-based~\cite{lin2025rgbdglass}), and the video-based VGSD-D dataset~\cite{Liu24videoglass}.

We report five standard evaluation metrics, including Intersection over Union (IoU$\uparrow$), F-measure ($F_{\beta}{\uparrow}$), Mean Absolute Error (MAE$\downarrow$), Balanced Error Rate (BER$\downarrow$), and Accuracy (ACC$\uparrow$).

\subsection{Comparisons to SOTA Methods}
{\bf Single-Image GSD Results.}
We first compare our method against six SOTA single-image GSD methods (GDNet, GSDNet, EBLNet, RFENet, GhostingNet, and GlassWizard) and three foundation models (SAM2, VGGT, and SAM3)
, on four standard benchmarks (GDD, GSD, HSO, and Trans10K-Stuff).
For a fair comparison, all competing methods are retrained under identical training settings.

%Tab.~\ref{tab:result} and~\ref{tab:trans10k} report the results, where our method demonstrates a consistent advantage in detection accuracy across all four datasets under all five evaluation metrics. 
\tao{The results reported in Tab.~\ref{tab:result} and~\ref{tab:trans10k} demonstrate that our method maintains a clear performance advantage across all evaluated datasets and metrics.}
%Notably, our method consistently outperforms the latest GhostingNet and GlassWizard, which verify the effectiveness of our method's grounding GSD task in visual geometry over the 2D ghosting effects cue~\cite{yan2025GhostingNet} and the large-scale 2D diffusion priors~\cite{li2025glasswizard}.
Notably, our method consistently outperforms the latest GhostingNet that relies on 2D ghosting cues and GlassWizard that leverages large-scale 2D diffusion priors, demonstrating the advantage of grounding GSD task in 3D visual geometry.
%\kk{In addition, Table~\ref{tab:trans10k} reports the results on the Trans10K-Stuff~\cite{xie2020Translab} dataset.}

%our method outperforms existing glass detection algorithms on both the GDD and GSD benchmark datasets. This indicates that the 3D information of the scene helps reveal the presence of glass, and the prior knowledge embedded in VGGT can effectively improve the accuracy of glass surface detection. Notably, our method shows significant improvement on the HSO dataset~\cite{yu2022pgs}. Compared to the second-best performing method GlassWizard~\cite{li2025glasswizard}, our method performs better by IoU: $4.14\uparrow$, $F_\beta$: $0.022\uparrow$, MAE: $0.024\downarrow$, BER: $2.71\downarrow$, and ACC: $0.032\uparrow$. This dataset contains a large number of indoor scenes with glass, which are more challenging due to the lack of prominent reflections. Compared to purely visual approaches, our model additionally leverages scene understanding to assist in detecting glass surfaces. Due to space limitations, we report the results on Trans10K-Stuff as well as the overall results on the four datasets in the appendix.
\begin{table}[t]
  \centering
  % \small
  \setlength{\tabcolsep}{3.5pt}
  \caption{Comparisons to multi-modal GSD methods, \ie, RGB-Thermal~\cite{huo2023glass} (upper) and RGB-Depth~\cite{lin2025rgbdglass} (lower) on their proposed datasets. Best results are in {\bf bold}.}
  % \vspace{-2mm}
  \begin{tabular}{lcccccc}
    \toprule
    Methods & Venue &IoU$\uparrow$ & F$_\beta$$\uparrow$ & MAE$\downarrow$ & BER$\downarrow$ & ACC$\uparrow$\\
    \midrule
    RGB-Thermal & TIP'23 & {93.80} & {0.965} & {0.027} & \bf{4.08} & - \\
    \rowcolor[gray]{.8}{\bf Ours} & - & \bf{95.71} & \bf{0.976} & \bf{0.023} & {4.19} & {\bf 0.970}\\
    \midrule
    RGB-Depth 
    & AAAI'25 & {74.20} & {0.853} & {0.043} & {9.30} & -\\
    \rowcolor[gray]{.8}{\bf Ours} & - & \bf{81.23} & \bf{0.873} & \bf{0.026} & \bf{7.05} & {\bf 0.880}\\
    \bottomrule
  \end{tabular}
  \label{tab:results-mutlimodel}
\vspace{-2mm}
\end{table}

\begin{table}[ht]
  \centering
  % \small
  \setlength{\tabcolsep}{3.8pt}
  \caption{Comparisons on the video GSD dataset VGSD-D~\cite{Liu24videoglass}. Best results are in {\bf bold}.}
  % \vspace{-2mm}
  \begin{tabular}{lcccccc}
    \toprule
    Methods & Venue &IoU$\uparrow$ & F$_\beta$$\uparrow$ & MAE$\downarrow$ & BER$\downarrow$ & ACC$\uparrow$\\
    \midrule
    % GSDNet &CVPR'21 &78.19 &86.46 &0.116 &11.1 &0.902\\
    % RFENet &IJCAI'23 &79.21 &88.60 &0.109 &10.5 &0.910\\
    VGSDNet &AAAI'24 &80.72 &0.885 &0.099 &9.60&0.898\\
    GhostingNet &TPAMI'25 &80.40 &0.888 &0.100 &9.30 &0.905\\
    GlassWizard &ICCV'25 &94.20 &- &0.031 &2.92 &-\\
    MVGDNet &AAAI'26 &86.54 &0.925 &0.064 &6.10 &0.935\\
   \rowcolor[gray]{.8}{\bf Ours} & - & \bf{97.36} & \bf{0.982} & \bf{0.016} & \bf{1.97} & \bf{0.986}\\
   \bottomrule
  \end{tabular}
  \label{tab:results-VGSD}
\vspace{-2mm}
\end{table}

{\bf Generalization Results.}
We evaluate our method's generalization capability by directly testing on two multi-modal GSD datasets (RGB-Thermal~\cite{huo2023glass} and RGB-Depth~\cite{lin2025rgbdglass}) and the video GSD dataset VGSD-D~\cite{Liu24videoglass}.
Tab.~\ref{tab:results-mutlimodel} reports the comparisons with multi-modal methods. Without requiring additional thermal or depth sensors, ~\lywcr{Our method achieves comparable performance to RGB-Thermal methods (Tab.~\ref{tab:results-mutlimodel} (top part)) while outperforming RGB-Depth methods (Tab.~\ref{tab:results-mutlimodel} (bottom part)).}
% our method achieves superior or comparable performance against the RGB-Thermal (Tab.~\ref{tab:results-mutlimodel} (top part)) and the RGB-Depth (Tab.~\ref{tab:results-mutlimodel} (bottom part)).
%
Tab.~\ref{tab:results-VGSD} further reports the comparisons on the VGSD-D. 
%GhostingNet tends to produce less satisfactory results when the ghosting effects cue may not be easily detected, while GlassWizard produces promising results by incorporating large-scale 2D diffusion priors. Our method outperforms GlassWizard consistently across metrics.
\tao{GhostingNet often degrades when ghosting cues are weak or ambiguous, whereas GlassWizard achieves stronger performance by leveraging large-scale 2D diffusion priors. Nevertheless, our method consistently outperforms GlassWizard across all evaluation metrics.}
%
% These results generally demonstrate the effectiveness of our learned geometry-aware GSD representations.

%We compared our method with the one proposed in the corresponding paper, and the results are shown in the Tab.~\ref{tab:results-mutlimodel}. Even without fine-tuning on these two datasets, our model can still achieve satisfactory results. We also verified the model's generalization capability on the video glass surface detection dataset VGSD-D~\cite{Liu24videoglass}, as shown in Tab~\ref{tab:results-VGSD}. Even when facing video data, our model has not shown any significant performance drop.
\begin{figure*}[t]
    \renewcommand{\tabcolsep}{0.8pt}
    \renewcommand\arraystretch{0.6}
    \renewcommand{\newsubwidth}{0.108}
    \centering
    \begin{tabular}{ccccccccc}
        \includegraphics[width=\newsubwidth\linewidth]{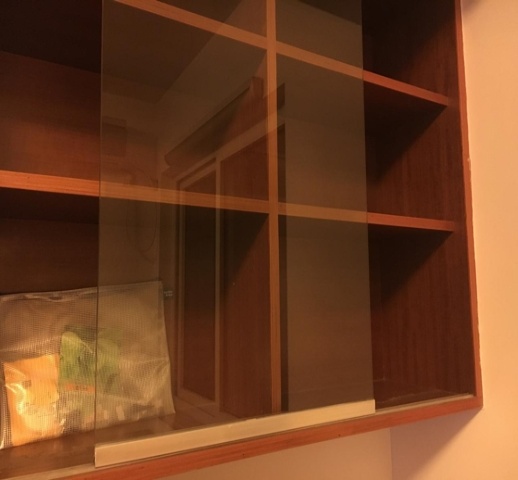}&
        
        \includegraphics[width=\newsubwidth\linewidth]{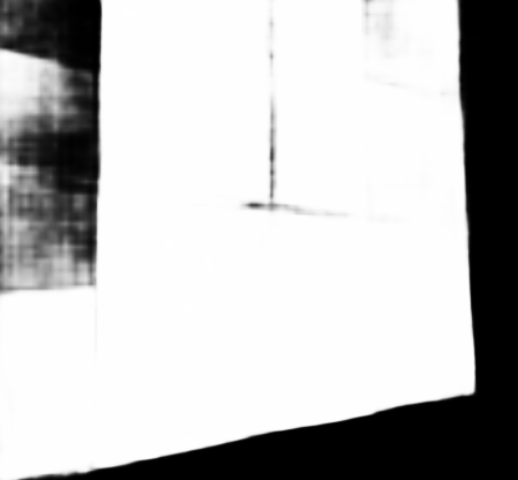}&
        \includegraphics[width=\newsubwidth\linewidth]{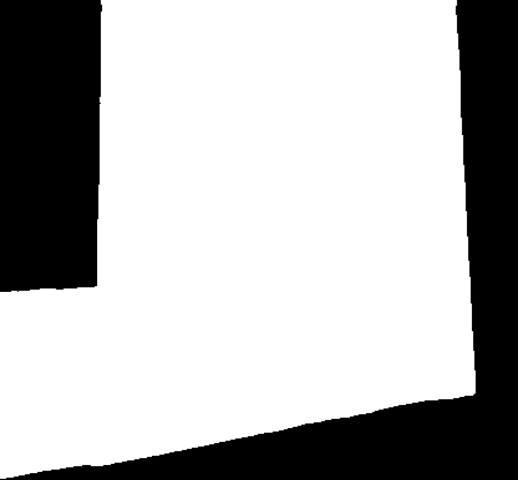}&
        \includegraphics[width=\newsubwidth\linewidth]{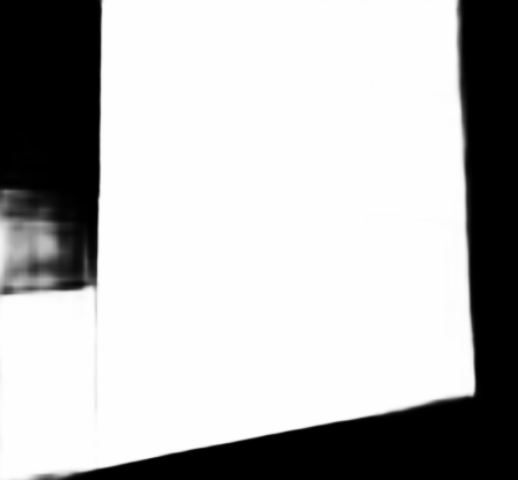}&
        \includegraphics[width=\newsubwidth\linewidth]{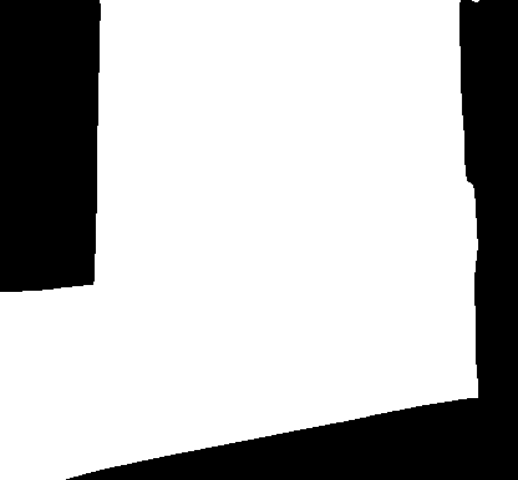}&
        \includegraphics[width=\newsubwidth\linewidth]{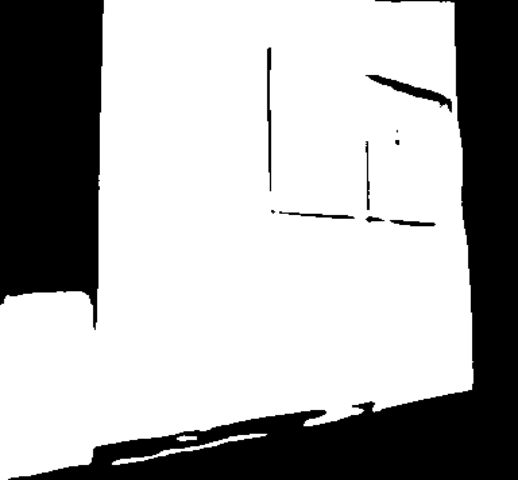}&
        \includegraphics[width=\newsubwidth\linewidth]{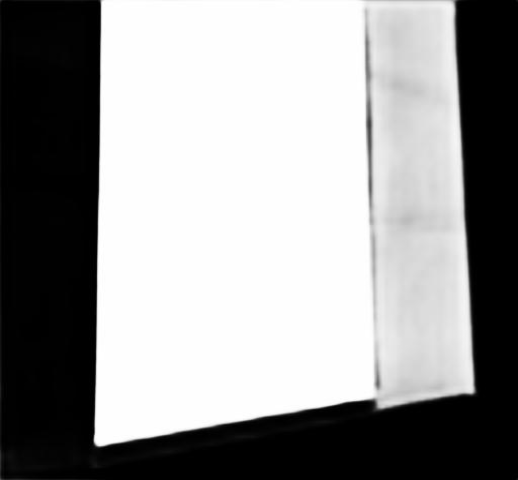}&
        \includegraphics[width=\newsubwidth\linewidth]{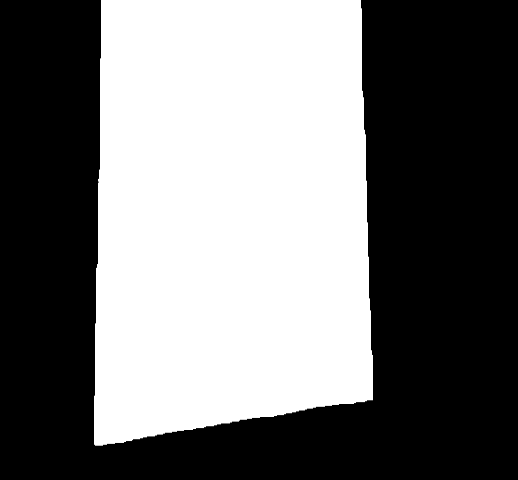}&
        \includegraphics[width=\newsubwidth\linewidth]{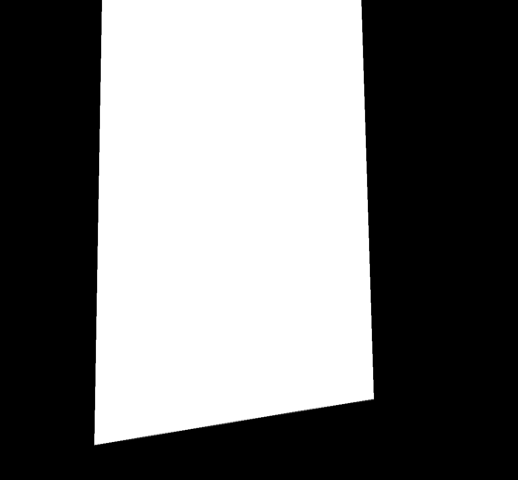}
        \\
        \includegraphics[width=\newsubwidth\linewidth]{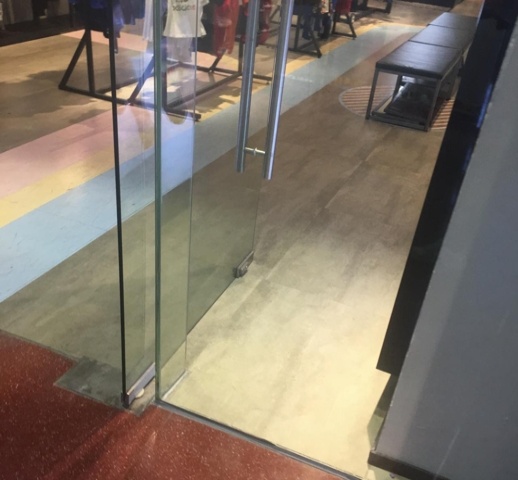}&
        \includegraphics[width=\newsubwidth\linewidth]{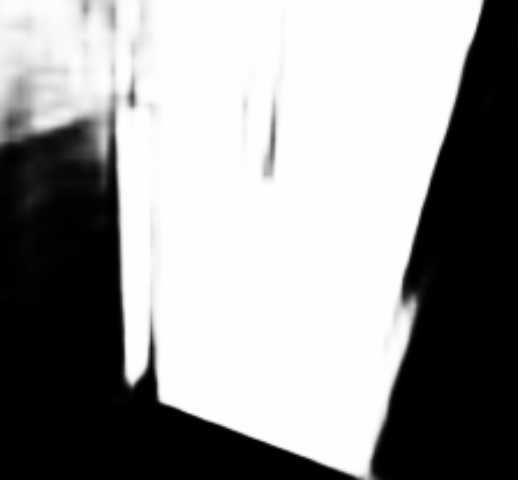}&
        \includegraphics[width=\newsubwidth\linewidth]{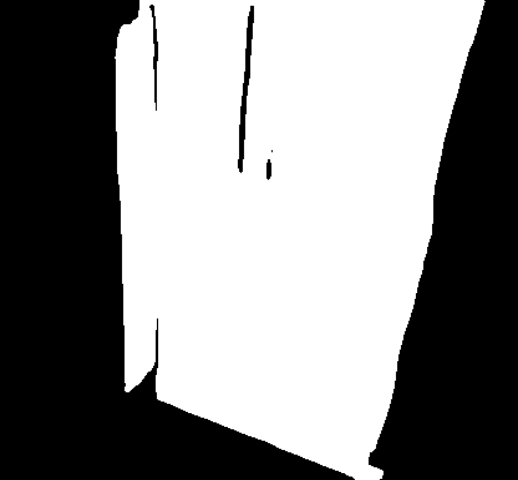}&
        \includegraphics[width=\newsubwidth\linewidth]{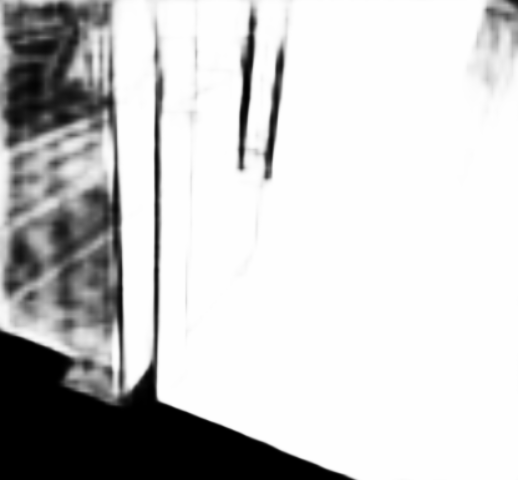}&
        \includegraphics[width=\newsubwidth\linewidth]{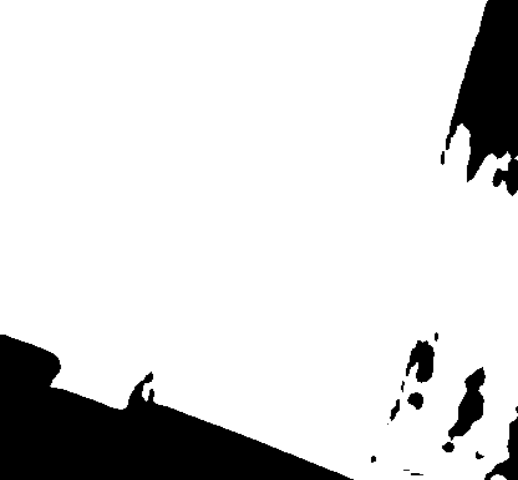}&
        \includegraphics[width=\newsubwidth\linewidth]{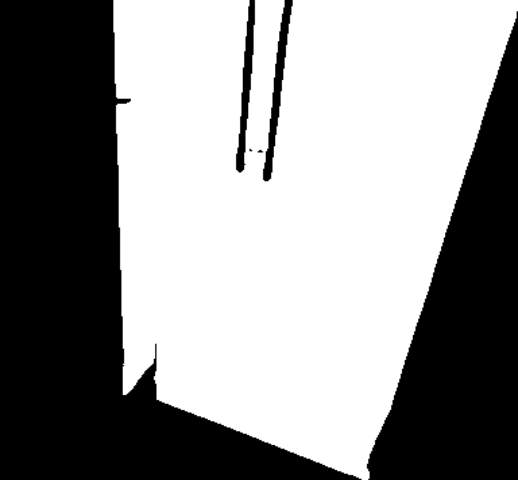}&
        \includegraphics[width=\newsubwidth\linewidth]{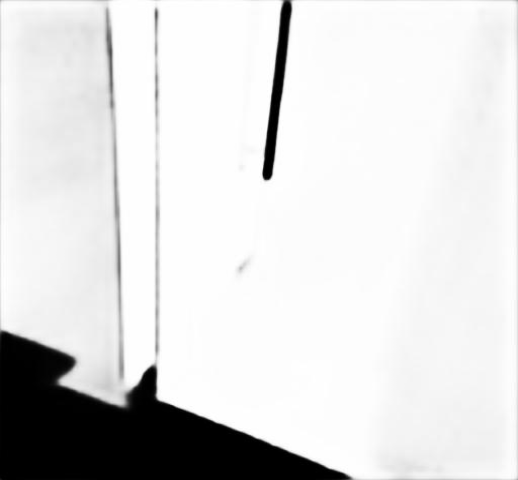}&
        \includegraphics[width=\newsubwidth\linewidth]{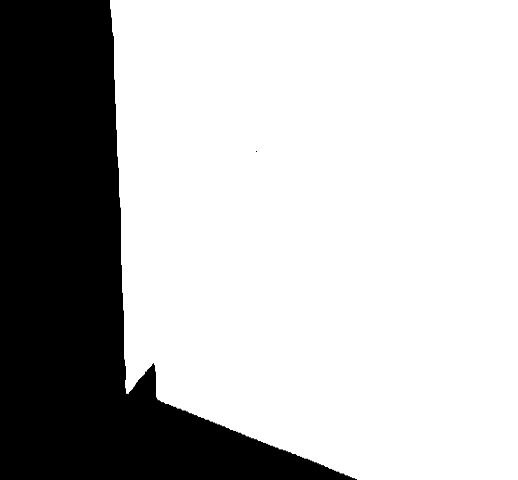}&
        \includegraphics[width=\newsubwidth\linewidth]{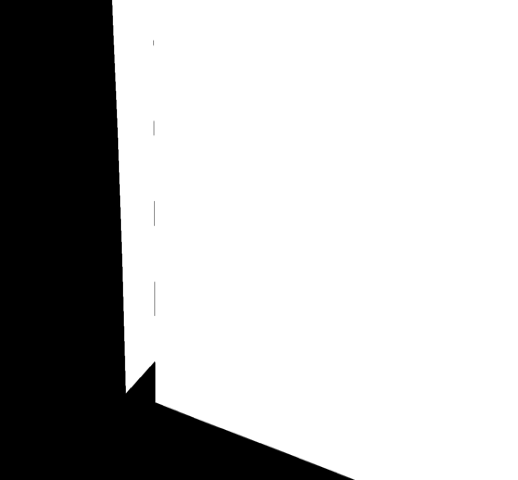}
        \\
        \includegraphics[width=\newsubwidth\linewidth]{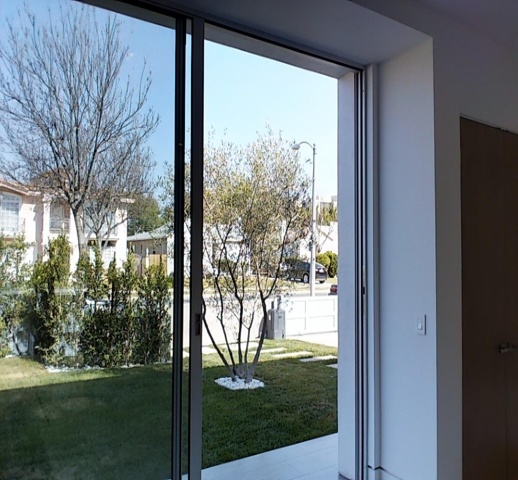}&
        \includegraphics[width=\newsubwidth\linewidth]{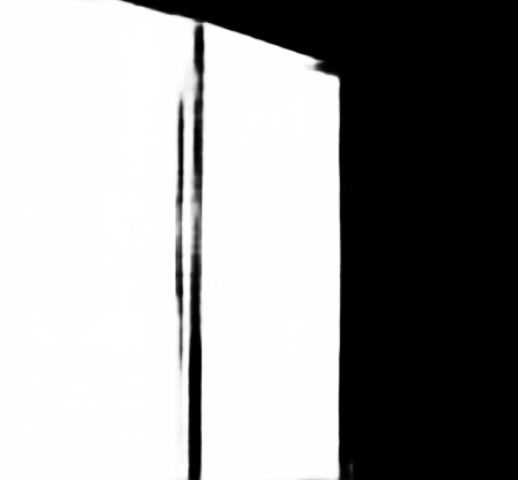}&
        \includegraphics[width=\newsubwidth\linewidth]{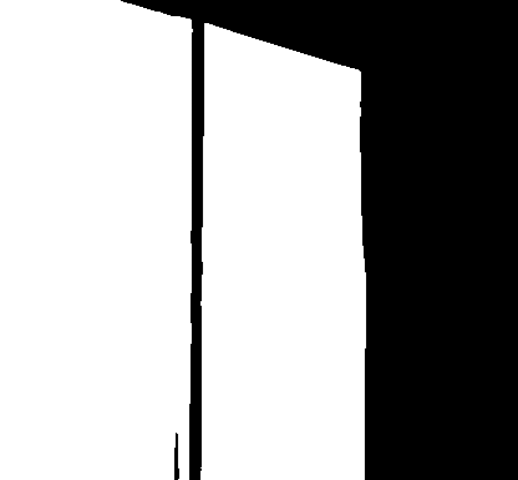}&
        \includegraphics[width=\newsubwidth\linewidth]{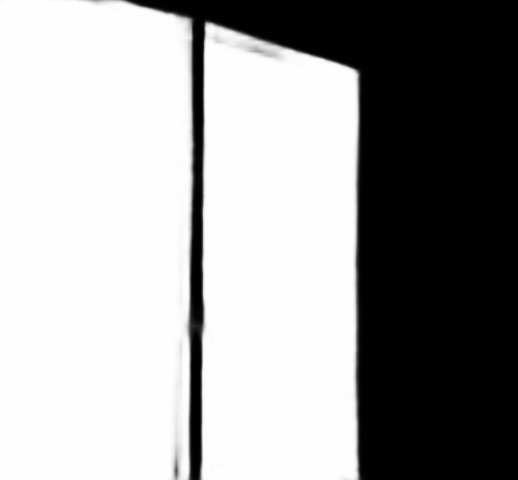}&
        \includegraphics[width=\newsubwidth\linewidth]{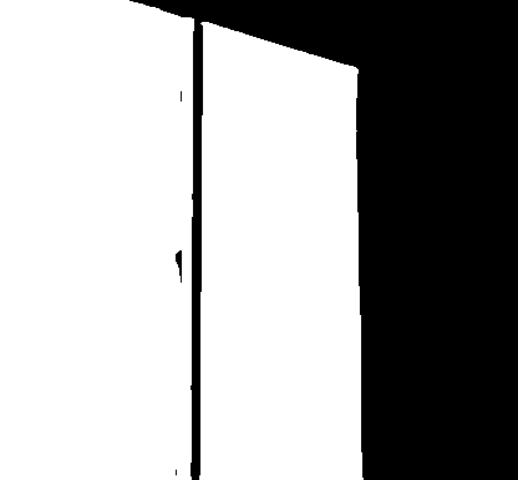}&
        \includegraphics[width=\newsubwidth\linewidth]{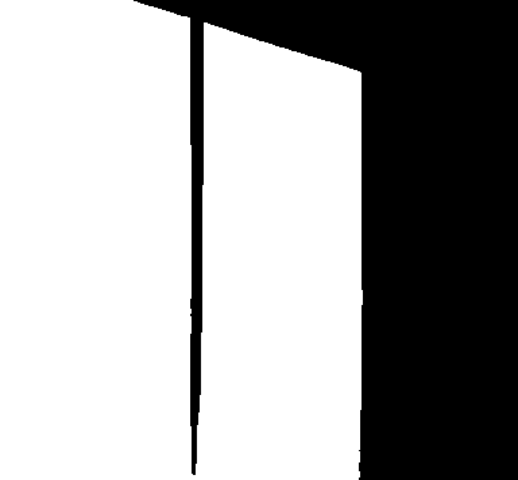}&
        \includegraphics[width=\newsubwidth\linewidth]{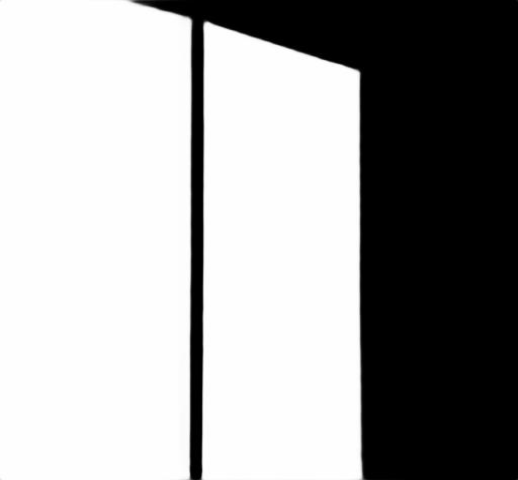}&
        \includegraphics[width=\newsubwidth\linewidth]{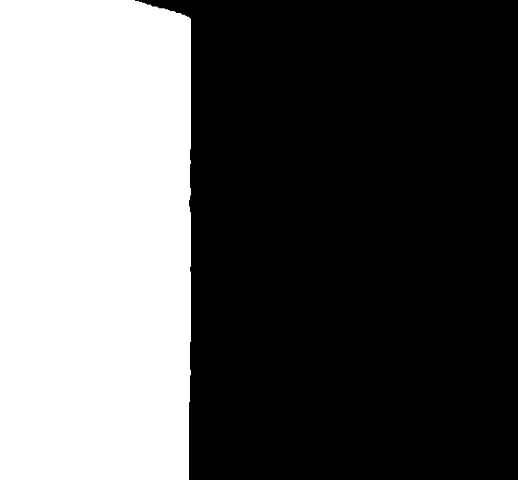}&
        \includegraphics[width=\newsubwidth\linewidth]{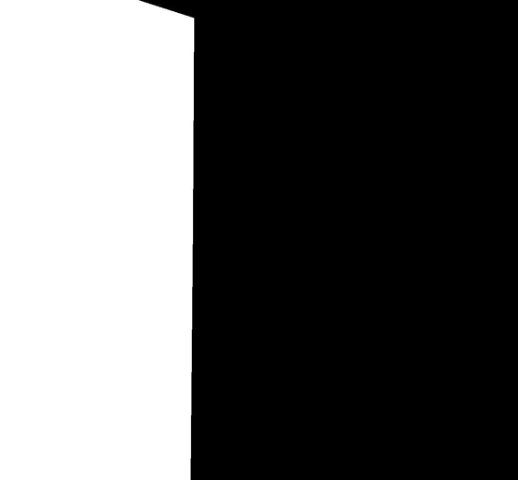}
        \\
        \includegraphics[width=\newsubwidth\linewidth]{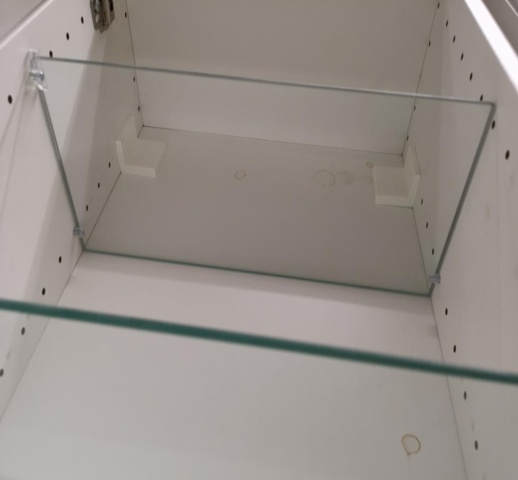}&
        \includegraphics[width=\newsubwidth\linewidth]{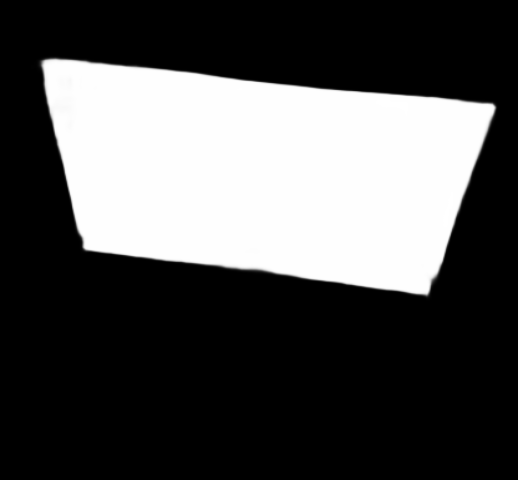}&
        \includegraphics[width=\newsubwidth\linewidth]{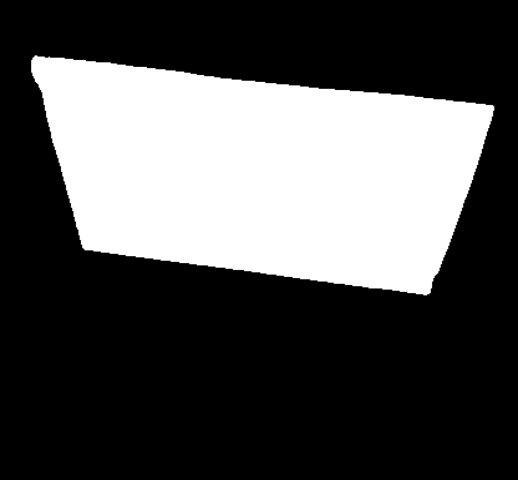}&
        \includegraphics[width=\newsubwidth\linewidth]{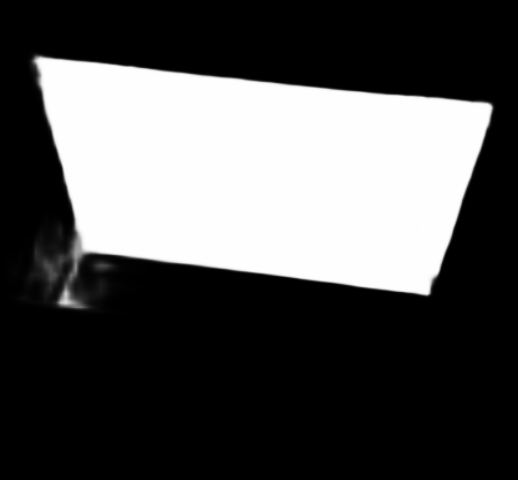}&
        \includegraphics[width=\newsubwidth\linewidth]{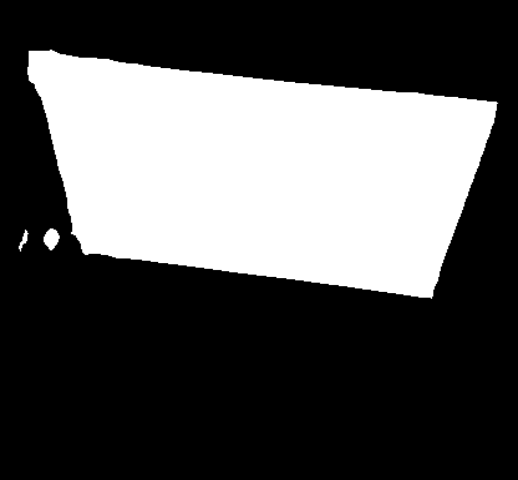}&
        \includegraphics[width=\newsubwidth\linewidth]{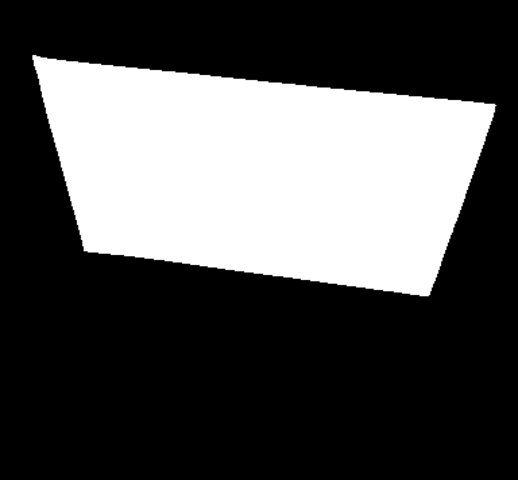}&
        \includegraphics[width=\newsubwidth\linewidth]{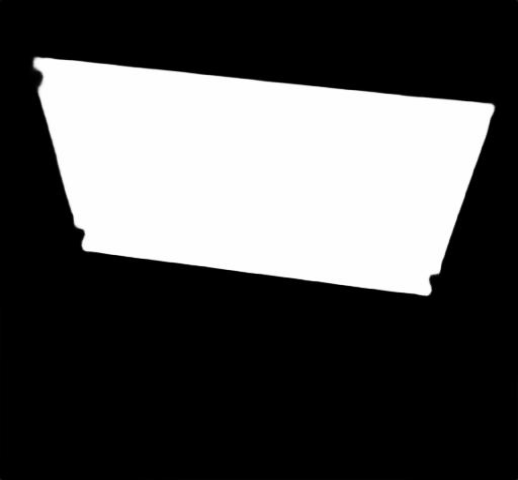}&
        \includegraphics[width=\newsubwidth\linewidth]{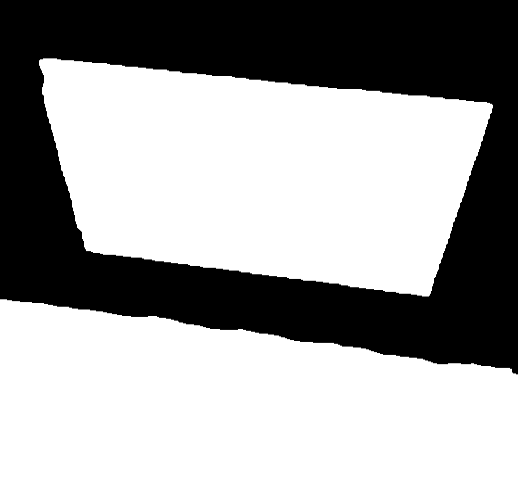}&
        \includegraphics[width=\newsubwidth\linewidth]{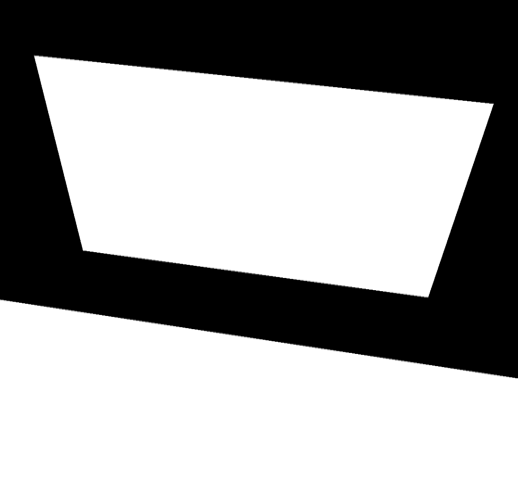}
        \\
        \includegraphics[width=\newsubwidth\linewidth]{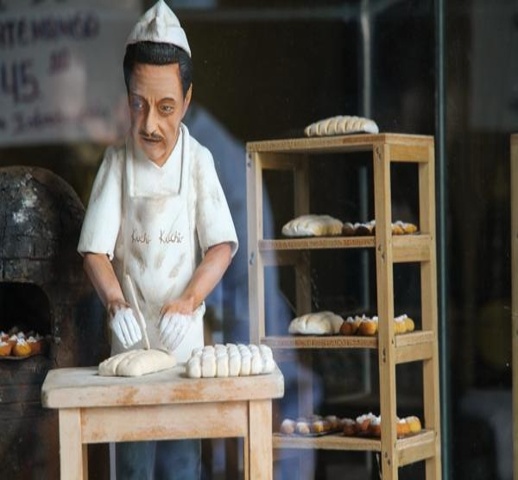}&
        \includegraphics[width=\newsubwidth\linewidth]{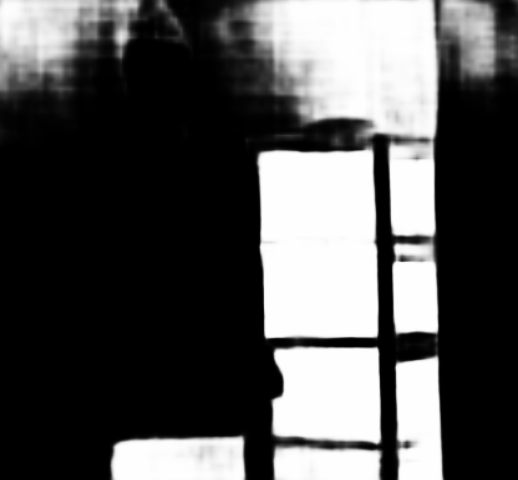}&
        \includegraphics[width=\newsubwidth\linewidth]{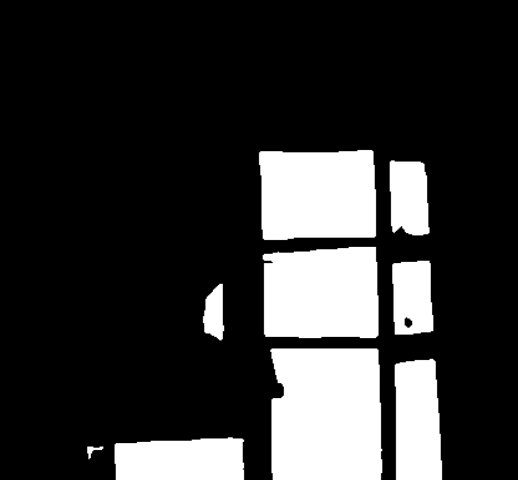}&
        \includegraphics[width=\newsubwidth\linewidth]{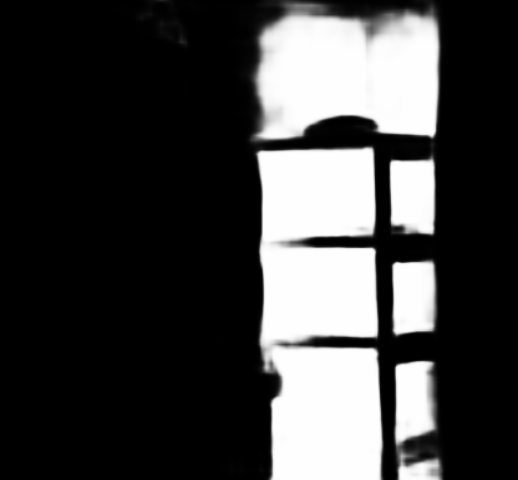}&
        \includegraphics[width=\newsubwidth\linewidth]{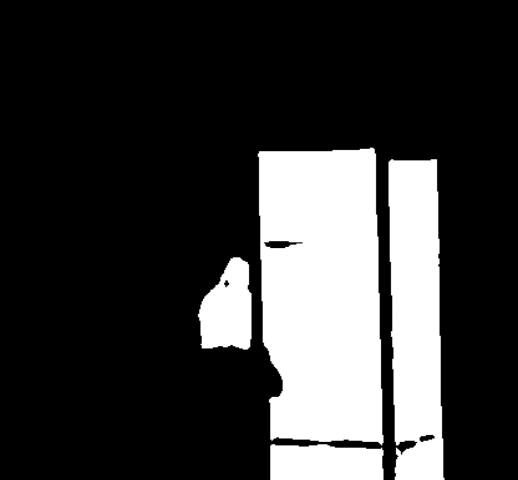}&
        \includegraphics[width=\newsubwidth\linewidth]{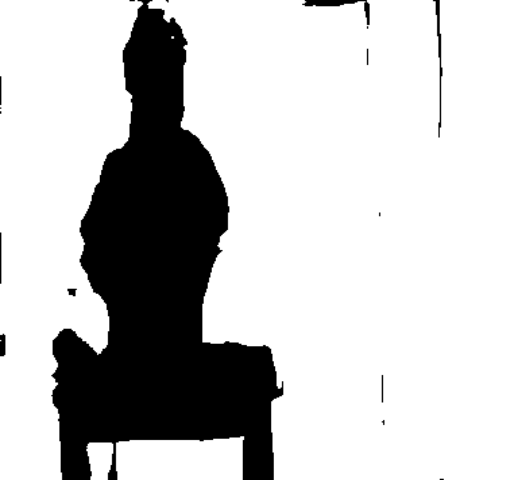}&
        \includegraphics[width=\newsubwidth\linewidth]{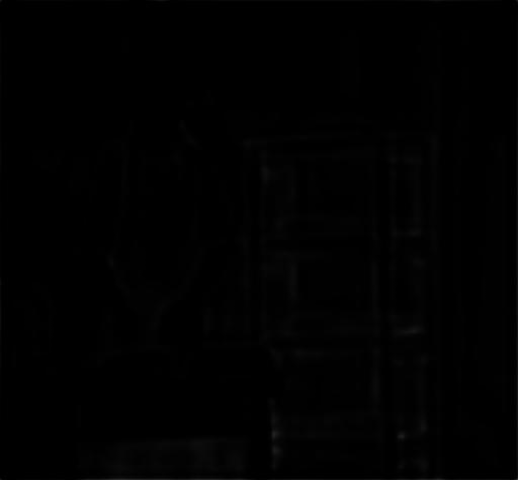}&
        \includegraphics[width=\newsubwidth\linewidth]{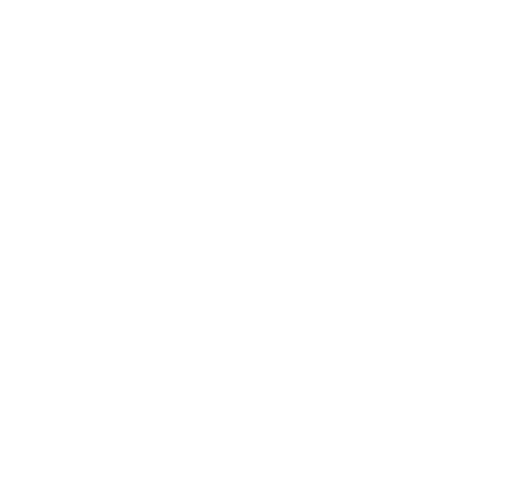}&
        \includegraphics[width=\newsubwidth\linewidth]{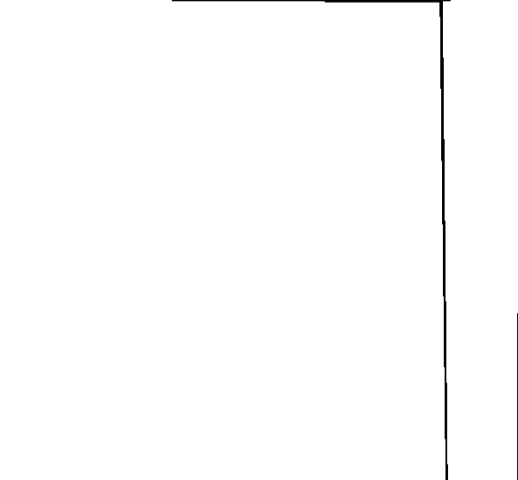}
        \\

        \includegraphics[width=\newsubwidth\linewidth]{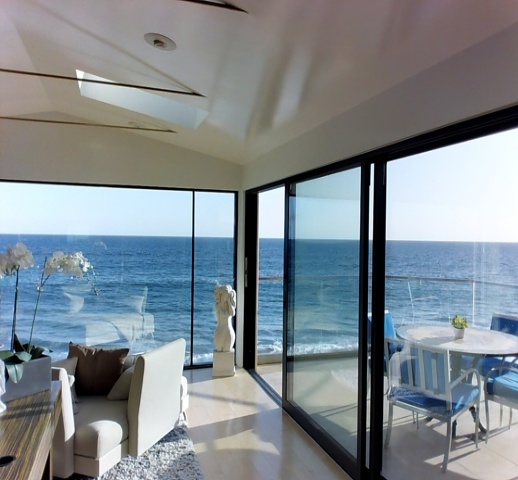}&
        \includegraphics[width=\newsubwidth\linewidth]{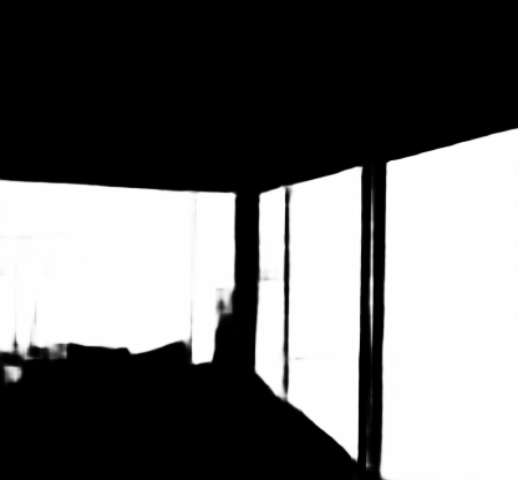}&
        \includegraphics[width=\newsubwidth\linewidth]{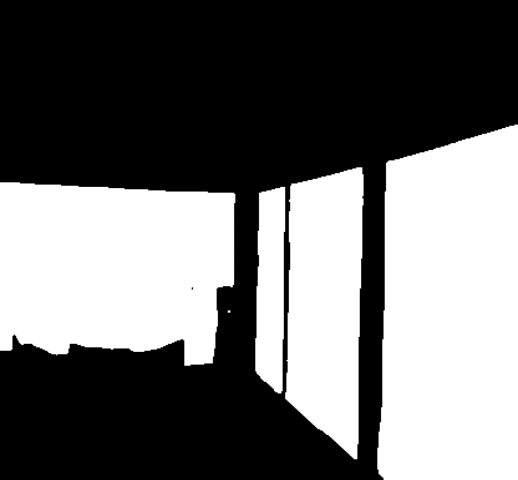}&
        \includegraphics[width=\newsubwidth\linewidth]{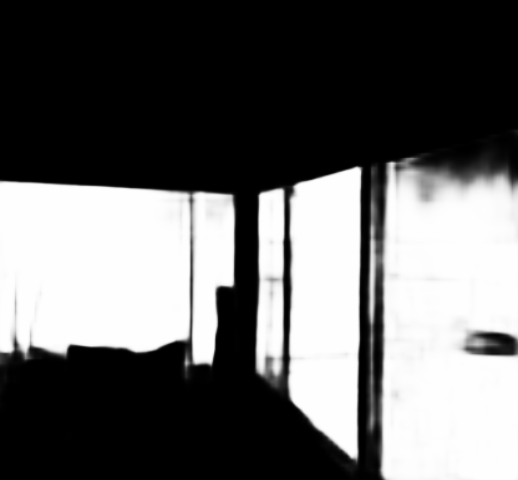}&
        \includegraphics[width=\newsubwidth\linewidth]{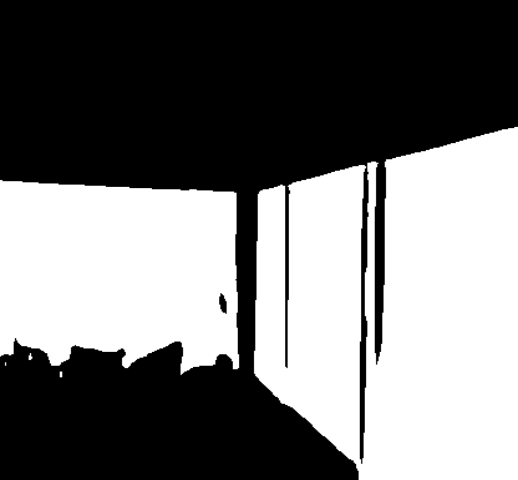}&
        \includegraphics[width=\newsubwidth\linewidth]{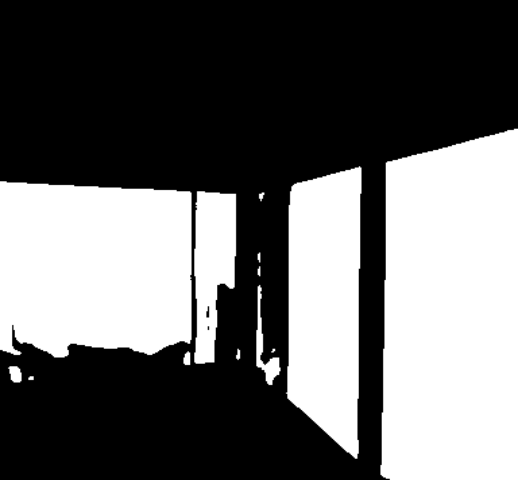}&
        \includegraphics[width=\newsubwidth\linewidth]{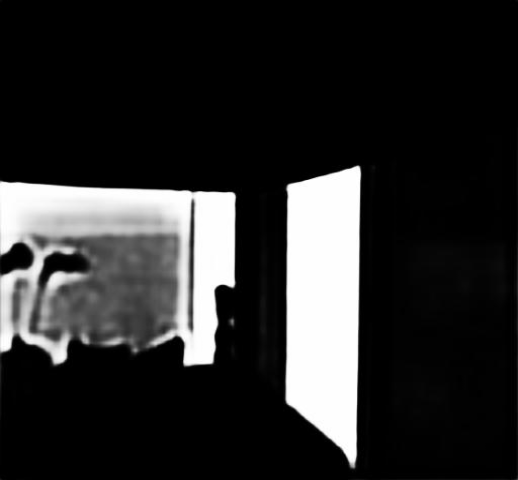}&
        \includegraphics[width=\newsubwidth\linewidth]{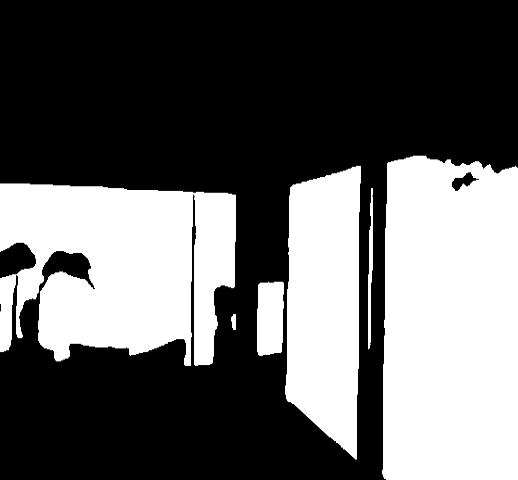}&
        \includegraphics[width=\newsubwidth\linewidth]{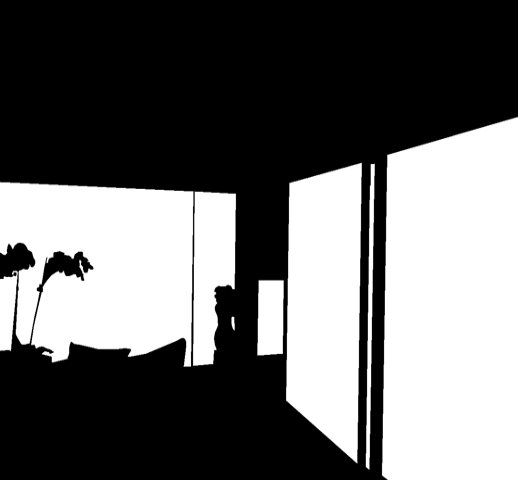}
        \\
        \small{RGB}&
        
        \small{GDNet}&
        \small{GSDNet}&
        \small{EBLNet}&
        \small{RFENet}&
        \small{Ghosting}&
        \small{GlassWizard}&
        %\small{Our Depth}&
        \small{Ours}&
        \small{GT}
    \end{tabular}
    \vspace{-3mm}
    \caption{Visual comparison between our method and single-image GSD methods, which highlights that 3D scene geometry information is vital for robust glass surface detection.}
    \label{fig:qualitative_comparison_single}
    \vspace{-3mm}
\end{figure*}

{\bf Visual Results.}
We provide visual comparisons with single-image-based GSD methods in Fig.~\ref{fig:qualitative_comparison_single}.
We can see that previous methods relying on 2D appearance cues may easily over- or under-detect glass regions that exhibit less distinct cues (\eg, reflections) and have similar visual patterns to those of non-glass regions, even with large-scale pre-trained 2D diffusion priors.
%
%In contrast, our method produces accurate glass surface segmentation results, effectively distinguishing glass regions from non-glass ones, as shown in the first four examples. 
\tao{In contrast, our method achieves accurate glass surface segmentation, clearly distinguishing glass from non-glass regions, as demonstrated in the first four examples.}
% In addition, we showcase two challenging examples: one where the glass region covers the entire image (the fifth example) and another where \lywcr{the glass regions} are partially occluded by objects (the last example). 
\lywcr{In addition, we showcase two challenging examples: a scene fully covered by glass (the fifth example) and a scene where the glass region is partially occluded by objects (the last example).}
While these two examples generally fail existing methods, our method can still accurately segment the glass surfaces and delineate their boundaries. These visual results highlight that 3D scene geometry information is vital for robust GSD.
%We show our depth predictions in the last third column, 
%Compared with competing methods, our model produces more accurate and consistent predictions. As shown in the second column, the visualized depth maps demonstrate that our approach effectively captures the underlying scene geometry. Moreover, our method performs well on challenging glass geometries, such as the glass sphere in the $2nd$ scene. Notably, the model also maintains reasonable accuracy in extreme cases, including the $4th$ scene, where the glass surface covers the entire image. These results indicate that 3D information can provide more robust information for glass surface detection.
\begin{figure*}[t]
	\renewcommand{\tabcolsep}{3pt}
	\renewcommand\arraystretch{0.6}
    \renewcommand{\newsubwidth}{0.13}
        \centering
        \small
            \begin{tabular}{ccccccc}
                % scene 2
                \includegraphics[width=\newsubwidth\linewidth]{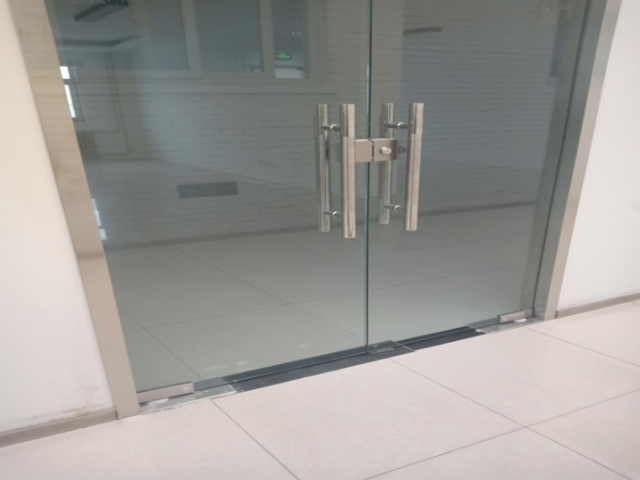}&
                \includegraphics[width=\newsubwidth\linewidth]{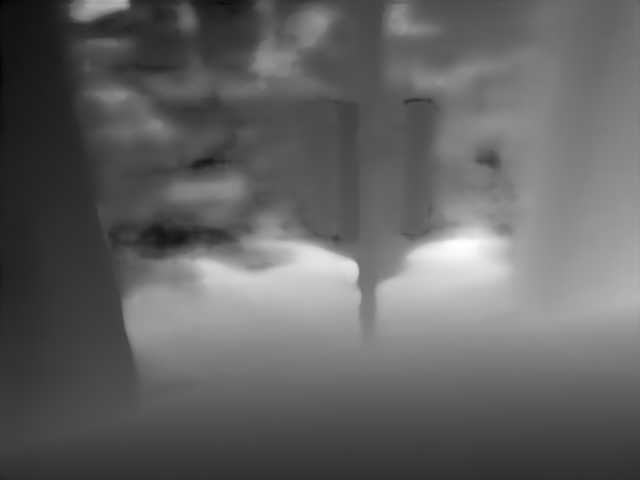}&
                \includegraphics[width=\newsubwidth\linewidth]{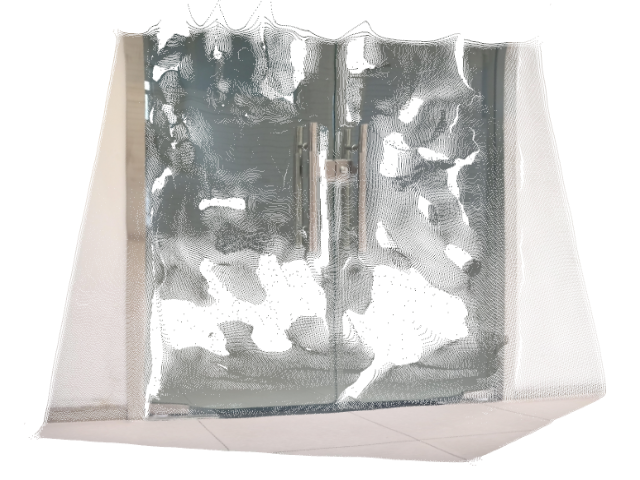}&
                \includegraphics[width=\newsubwidth\linewidth]{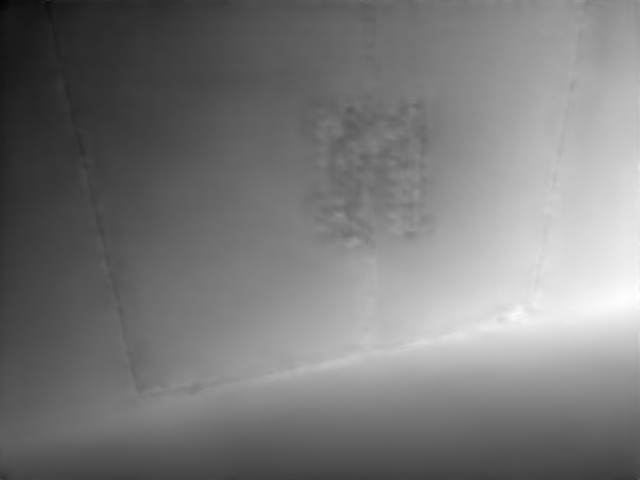}&
                \includegraphics[width=\newsubwidth\linewidth]{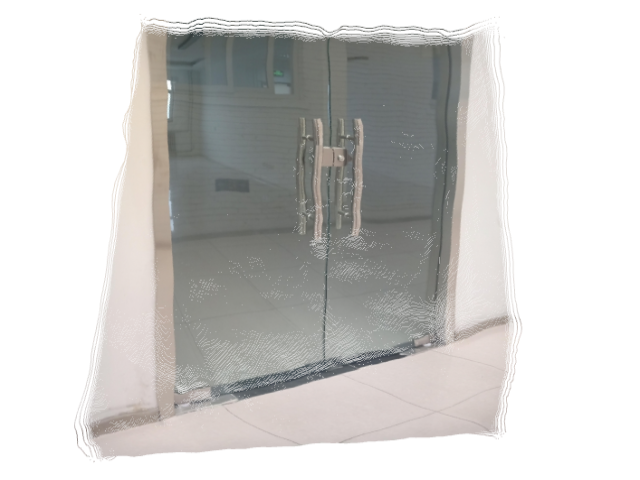}&
                \includegraphics[width=\newsubwidth\linewidth]{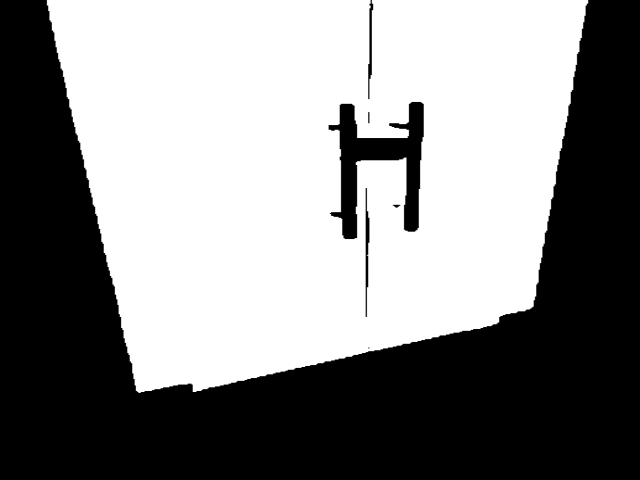}&
                \includegraphics[width=\newsubwidth\linewidth]{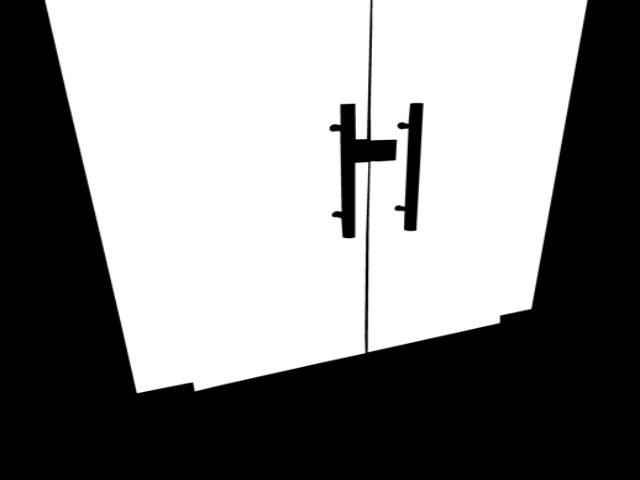}
                \\
                \includegraphics[width=\newsubwidth\linewidth]{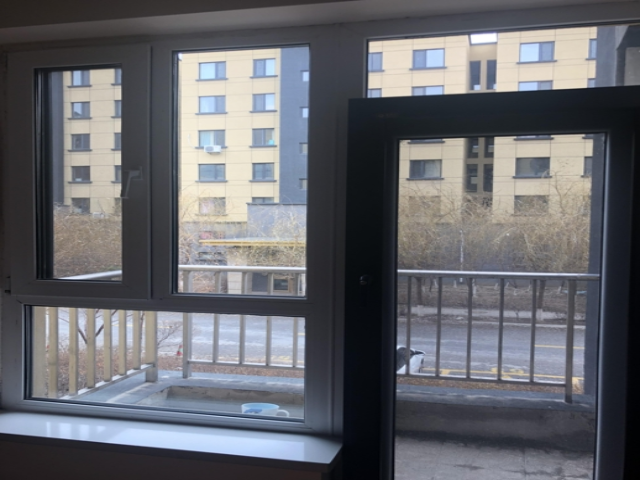}&
                \includegraphics[width=\newsubwidth\linewidth]{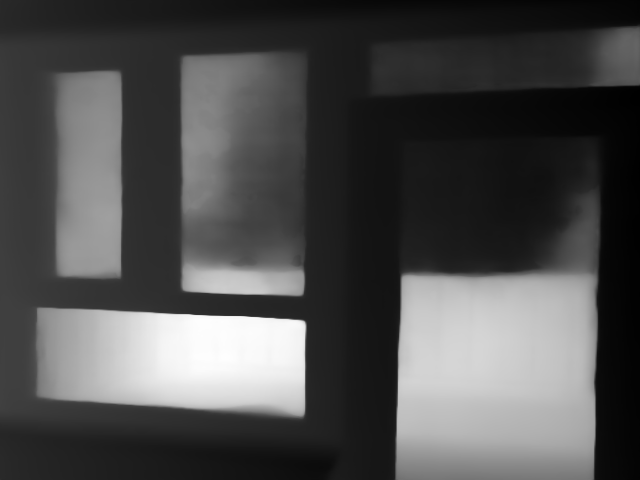}&
                \includegraphics[width=\newsubwidth\linewidth]{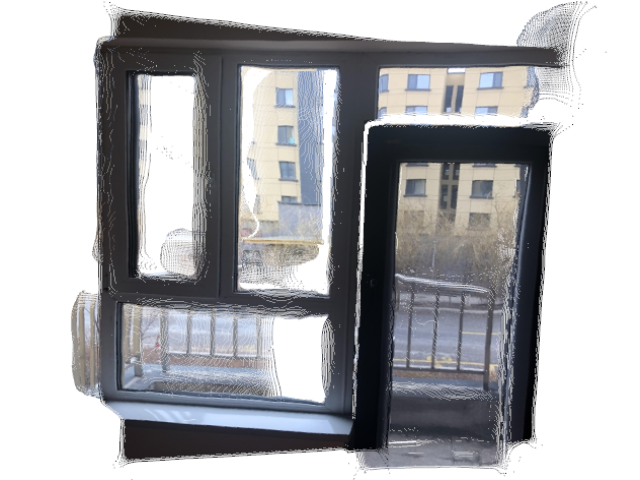}&
                \includegraphics[width=\newsubwidth\linewidth]{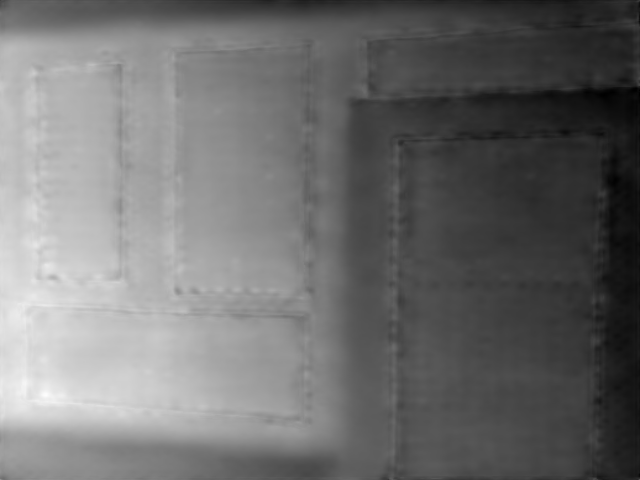}&
                \includegraphics[width=\newsubwidth\linewidth]{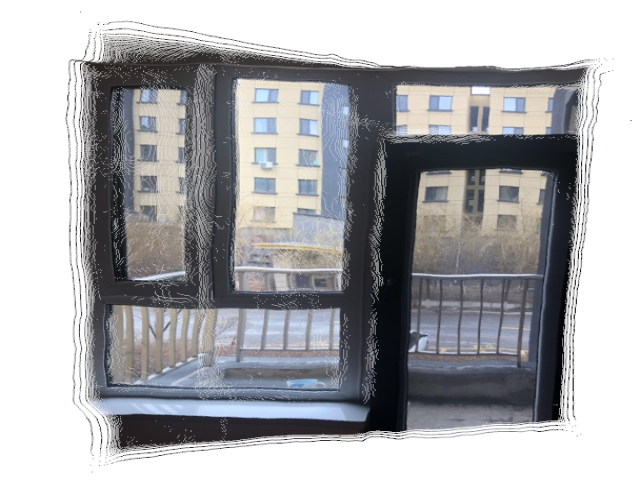}&
                \includegraphics[width=\newsubwidth\linewidth]{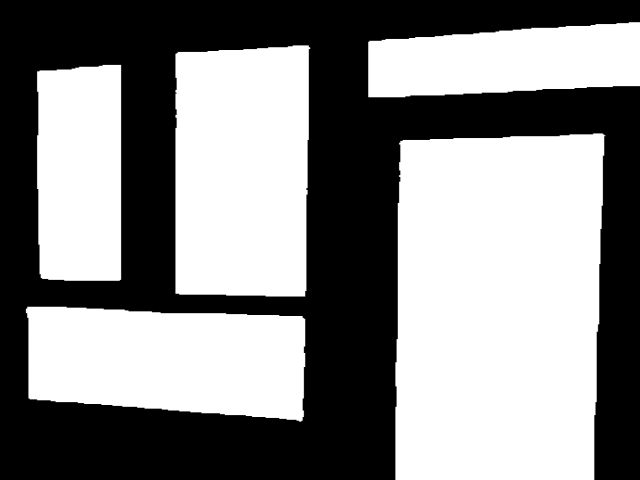}&
                \includegraphics[width=\newsubwidth\linewidth]{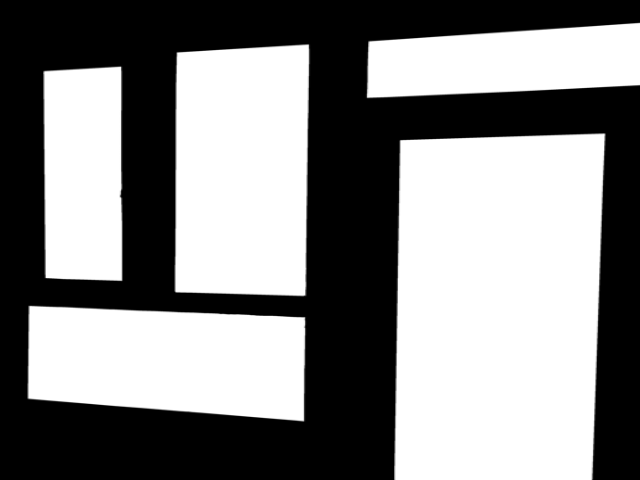}
                \\
                
                \includegraphics[width=\newsubwidth\linewidth]{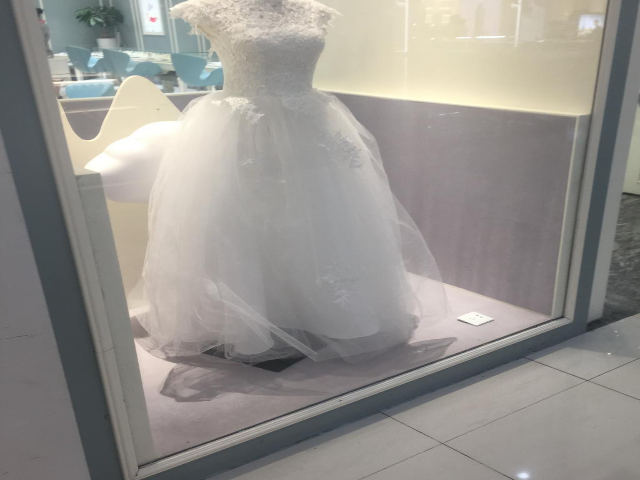}&
                \includegraphics[width=\newsubwidth\linewidth]{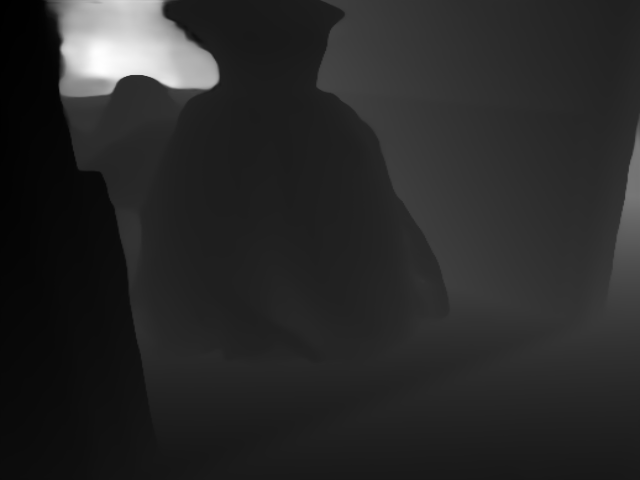}&
                \includegraphics[width=\newsubwidth\linewidth]{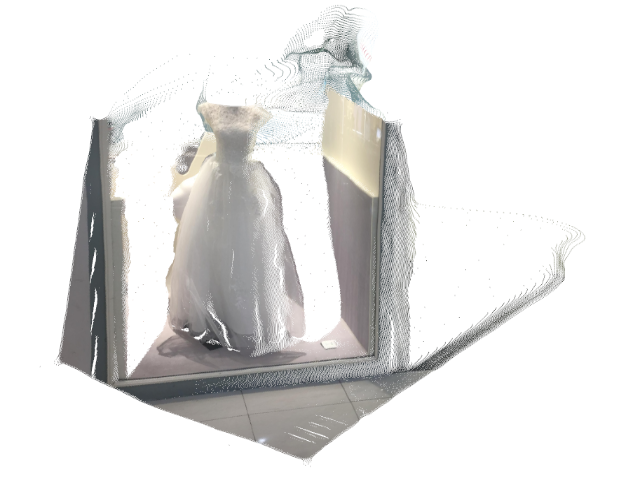}&
                \includegraphics[width=\newsubwidth\linewidth]{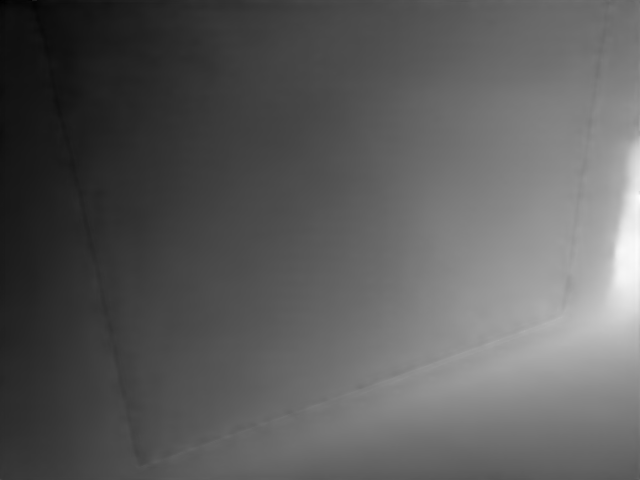}&
                \includegraphics[width=\newsubwidth\linewidth]{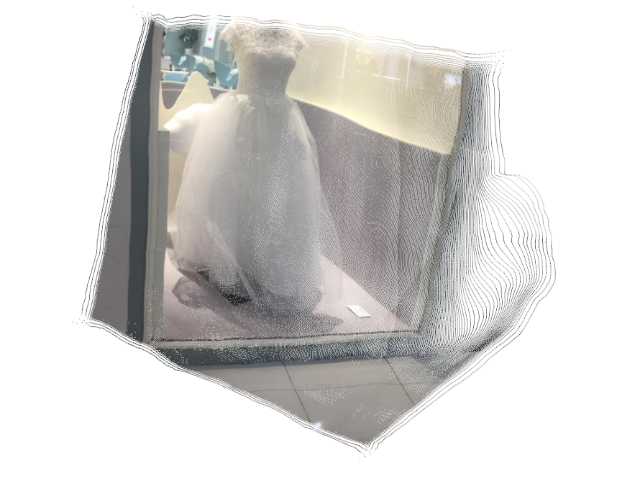}&
                \includegraphics[width=\newsubwidth\linewidth]{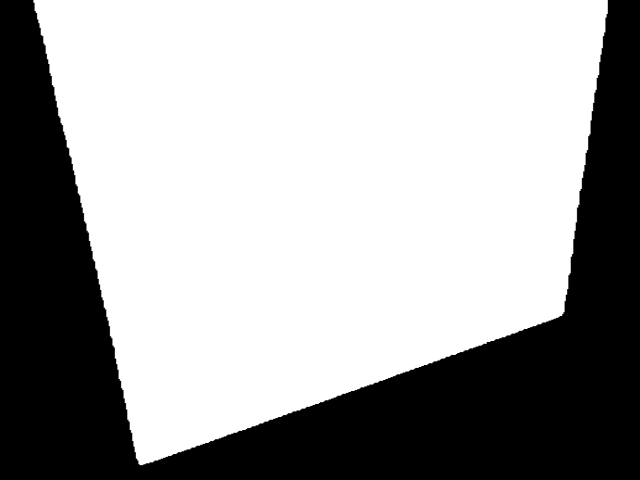}&
                \includegraphics[width=\newsubwidth\linewidth]{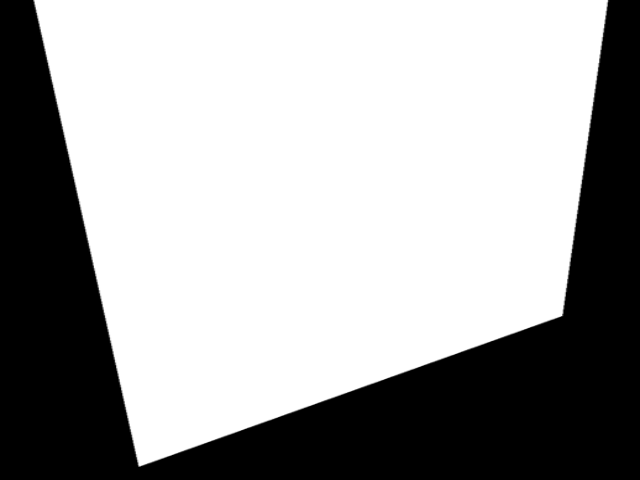}
                \\
                
                % annotation
                RGB Input&
                VGGT Depth&
                VGGT Point Cloud&
                Our Depth&
                Our Point Cloud&
                % \footnotesize Ghosting&
                %\footnotesize Glasswizard&
                Our GSD&
                GT GSD
            \end{tabular}
    \vspace{-3mm}
    \caption{Visual comparison between VGGT~\cite{wang2025vggt} and our method in monocular depth estimation and reconstruction of 3D scenes with glass surfaces.}
    \label{fig:3Drecon}
    \vspace{-3mm}
\end{figure*}

{\bf Application.} 
\lywre{To further demonstrate the effectiveness of our geometry-aware representation, we evaluate its impact on 3D reconstruction in scenes containing glass surfaces.
As shown in Fig.~\ref{fig:3Drecon}, the original VGGT~\cite{wang2025vggt} tends to reconstruct the 3D structure of objects behind transparent surfaces, often neglecting the physical presence of glass. In contrast, our method explicitly incorporates planar geometric priors of glass surfaces, leading to more accurate reconstruction in glass regions.
Exploring more expressive representations for multi-layered point cloud reconstruction remains an interesting direction for future work.}

% \begin{table}[h]
%   \centering
%   \setlength{\tabcolsep}{1pt}
%   \caption{Quantitative results on glass-free scenes from the RGB-T dataset.}
%   %\vspace{-2mm}
%   \begin{tabular}{@{}llccccc@{}}
%     \toprule
%     Methods & Venue &IoU*↑ & FPR↓ & MAE↓\\
%     \midrule
%     RGB-T GSD~\cite{huo2023glass} & TIP'23 & {99.73} & \bf{0.08\%} & {0.003}\\
%     Ours & - & \bf{99.90} & {0.11\%} & \bf{0.0013}\\
%     \bottomrule
%   \end{tabular}
%   \label{tab:results-noglass}
% \vspace{-3mm}
% \end{table}

\begin{table}[t]
  \centering
  \setlength{\tabcolsep}{6pt}
  % \small
  \caption{Ablation results of our method.}
  \begin{tabular}{lccccc}
    \toprule
    Methods & IoU$\uparrow$ & F$_\beta$$\uparrow$  & MAE$\downarrow$ & BER$\downarrow$ & ACC$\uparrow$\\
    \midrule
    w/o LoRA & 88.27 & 0.931 & 0.046 & 4.05 & 0.981\\ 
    w/o FSAM & 92.78 &  0.964 & 0.027 & 2.29 & 0.988\\
    w/o GeGB & 92.67 & 0.963  & 0.027 & 2.31 & 0.988\\
    w/o FFT    & 92.82 & 0.967 & 0.027 & 2.37 & 0.986\\
    $F_{in}\rightarrow F_{fre}$   & 93.06 & 0.967 & 0.026 & {\bf2.12} & {\bf0.991}\\
    w/o Depth  & 92.86 &  0.967 & 0.026 & 2.28 & 0.988 \\
    w/o Point  & 92.70 &  0.965 & 0.027 & 2.25 & 0.990 \\
    \midrule
    w/o P\&D-C & 93.04 & 0.965 &0.028 &2.36 &0.985 \\
    w/o P-C& 92.84 & 0.963 &0.029 &2.43 &0.984 \\
    w/o D-C& 92.76 & 0.962 &0.029 &2.40 &0.987 \\
    % w P\&D & - & - & - & - & - \\
    w/o $G^c$&92.98&0.968&0.026&2.24&0.988\\
    w/o $\mathcal{L}_{b}$  & 92.75 &  0.967 & 0.026 & 2.43 & 0.983 \\
    % \lywcr{w/o 3D loss} & 92.75 &  0.967 & 0.026 & 2.43 & 0.983 \\
    \midrule
    %Baseline & 85.77 & 0.913 & 0.055 & 4.73 & 0.981 \\
    %Baseline + LoRA  & 92.48 &  0.958 & 0.028 & 2.47 & {\bf 0.991}\\
    %VGGT + LoRA & 92.85 & 0.966 & 0.027 & 2.29 & 0.988 \\
    VGGT$\rightarrow$SAM3 & 92.34 & 0.960 & 0.030 & 2.51 & 0.985 \\
    \midrule
    \rowcolor[gray]{.8}{\bf Ours} & {\bf 93.24} & {\bf 0.969} & {\bf 0.025} & {2.15} & {0.989} \\
    \bottomrule
  \end{tabular}
  \label{tab:ablation_study_module}
  \vspace{-3mm}
\end{table}

\subsection{Ablation Studies}
We conduct ablation studies based on the Union glass test set (which combines the test sets from GDD, GSD, Trans10K-Stuff, and HSO) to study their broader impact on the GSD task instead of individual datasets.
Tab.~\ref{tab:ablation_study_module} reports the results.

\begin{figure}[t]
	\renewcommand{\tabcolsep}{0.5pt}
	\renewcommand\arraystretch{0.5}
    \renewcommand{\newsubwidth}{0.135}
        \centering
        \small
            \begin{tabular}{ccccccc}
                % scene 2
                \includegraphics[width=\newsubwidth\linewidth]{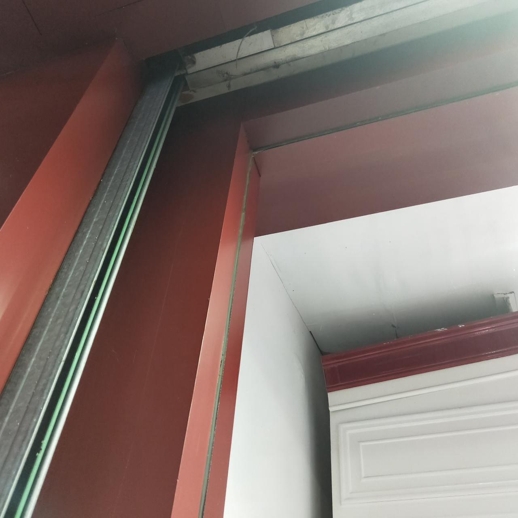}&
                \includegraphics[width=\newsubwidth\linewidth]{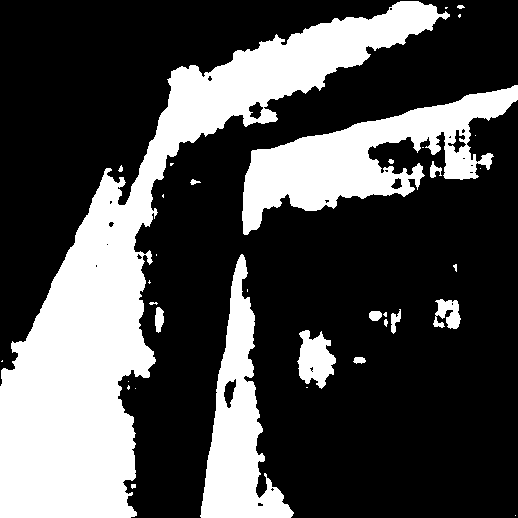}&
                \includegraphics[width=\newsubwidth\linewidth]{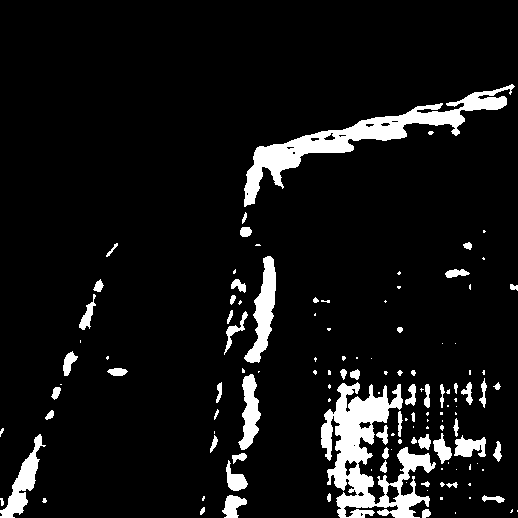}&\includegraphics[width=\newsubwidth\linewidth]{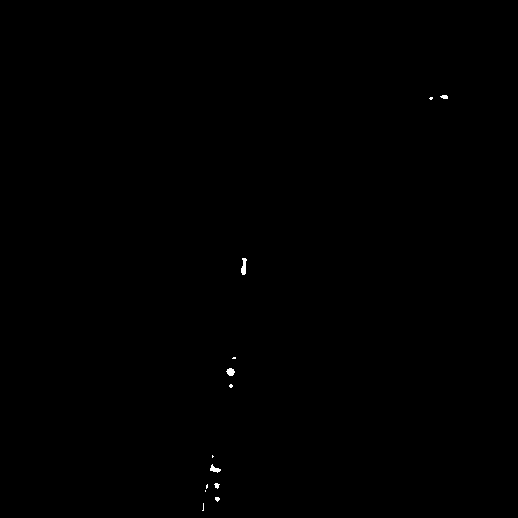}&
                \includegraphics[width=\newsubwidth\linewidth]{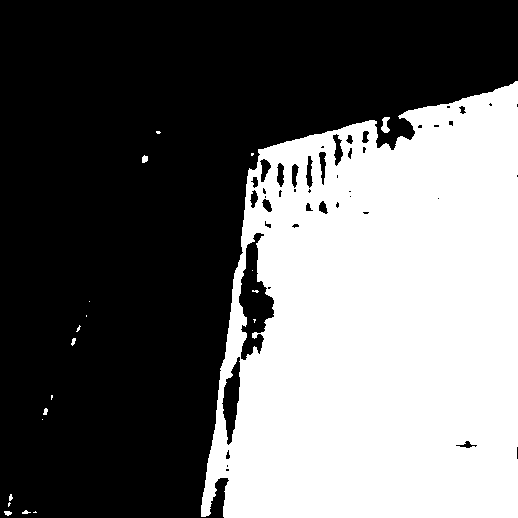}&
                \includegraphics[width=\newsubwidth\linewidth]{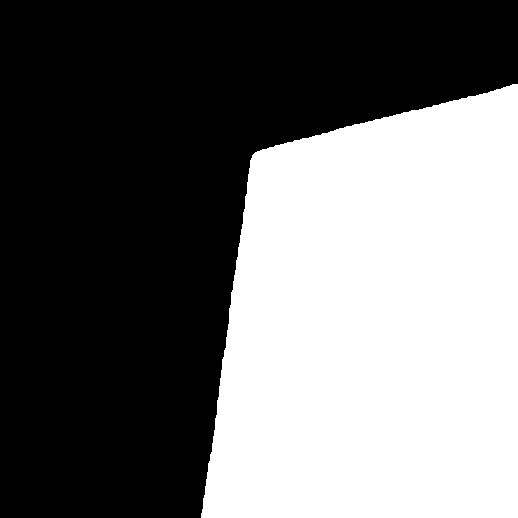}&
                \includegraphics[width=\newsubwidth\linewidth]{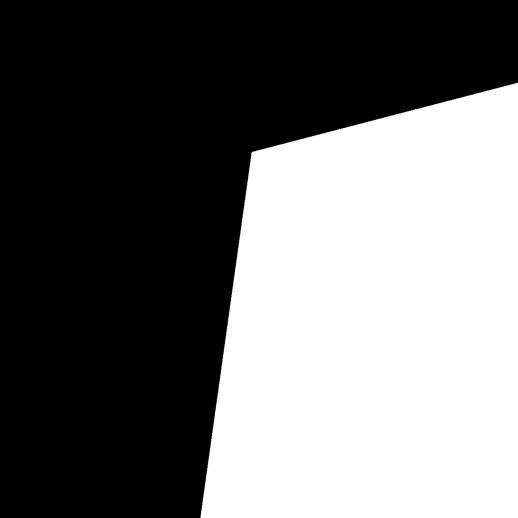}
                \\
                \includegraphics[width=\newsubwidth\linewidth]{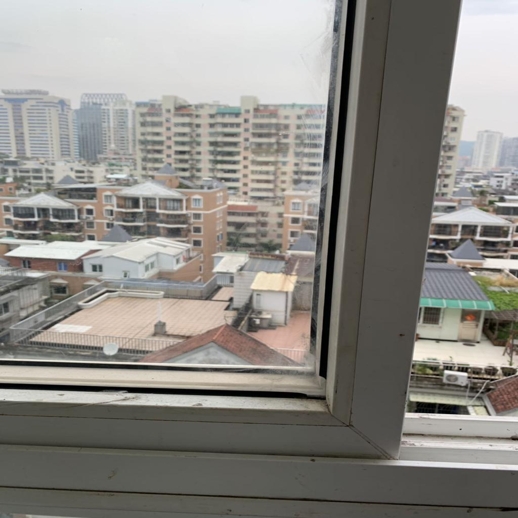}&
                \includegraphics[width=\newsubwidth\linewidth]{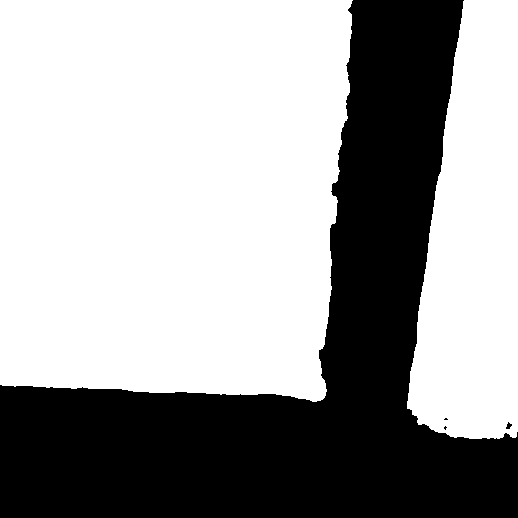}&
                \includegraphics[width=\newsubwidth\linewidth]{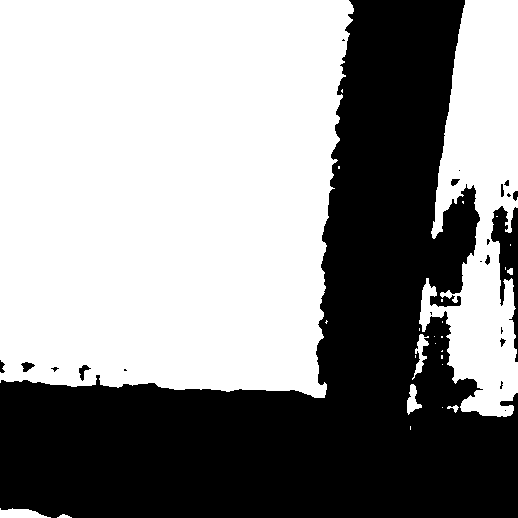}&\includegraphics[width=\newsubwidth\linewidth]{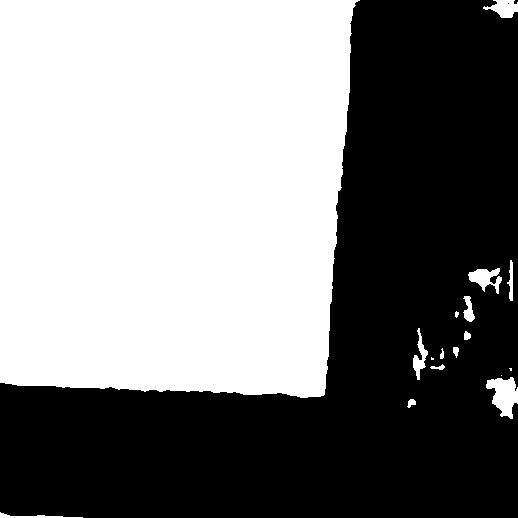}&
                \includegraphics[width=\newsubwidth\linewidth]{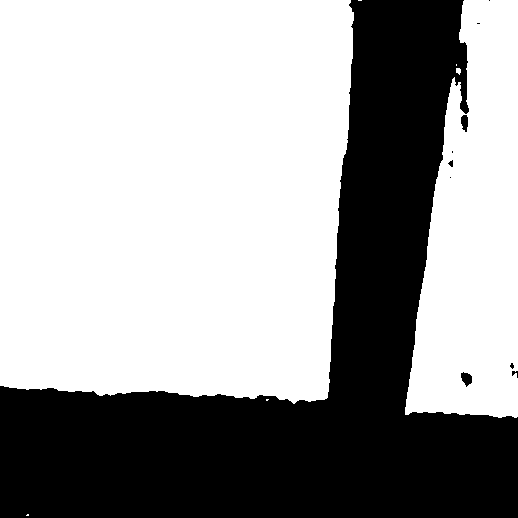}&
                \includegraphics[width=\newsubwidth\linewidth]{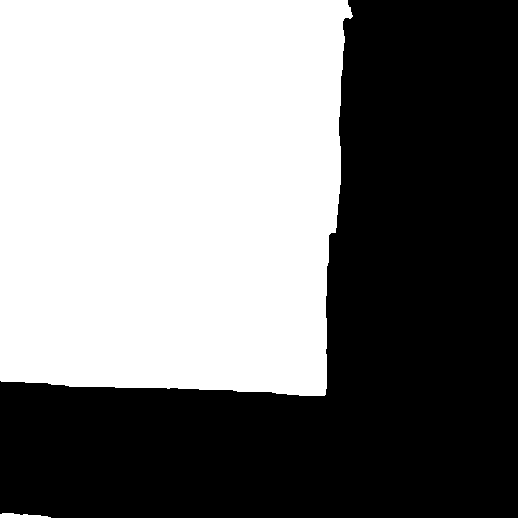}&
                \includegraphics[width=\newsubwidth\linewidth]{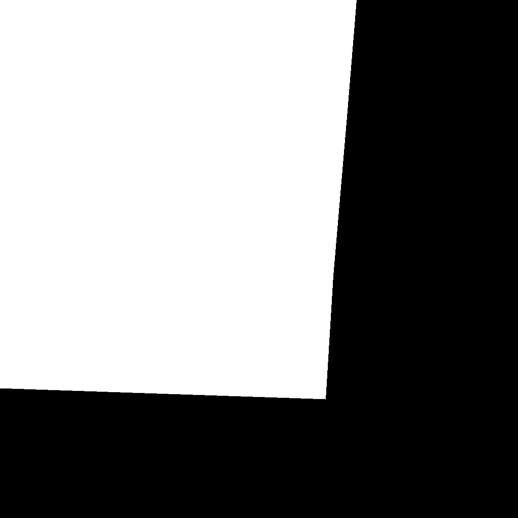}
                \\
                
                % annotation
                RGB&
                Baseline&
                \makecell[c]{w/o \\ LORA}&
                \makecell[c]{w/o \\ FSAM}&
                \makecell[c]{w/o \\ GeGB}&
                % \footnotesize Ghosting&
                %\footnotesize Glasswizard&
                Ours&
                GT
            \end{tabular}
    \vspace{-3mm}
    \caption{The visual comparison of different ablated models.}
    \label{fig:adlation_fig}
    \vspace{-4mm}
\end{figure}

{\bf Model Designs.}
As shown in Tab.~\ref{tab:ablation_study_module} (top part), we first analyze network components by removing each component individually and retraining the ablated model, denoted as ``w/o LoRA'', ``w/o FSAM'', and ``w/o GeGB''. We also remove Eq.~\ref{eq:fft} in FSAM (which then becomes a standard \tao{spatial-domain} self-attention mechanism) and denote it as ``w/o FFT''. \lyw{We also apply the self-attention mechanism completely in the frequency domain, \ie, replacing $F_{in}$ with $F_{fre}$ in Eq.~\ref{eq:attention} and denote it as ``$F_{in}\rightarrow F_{fre}$''.} In addition, we also remove the depth and point paths individually from our GeGB, denoted as ``w/o Depth'' and ``w/o Point'', respectively. These ablated models tend to yield sub-optimal performance. Notably, while ``$F_{in}\rightarrow F_{fre}$'' produces slightly better BER and accuracy results, it leads to a relatively larger IoU performance drop, which shows that completely relying on frequency information tends to over-detect glass regions.
These results generally verify the effectiveness of our model designs. \lywre{Visual comparisons are shown in Fig.~\ref{fig:adlation_fig}. FSAM captures semantic cues for glass localization, while GeGB leverages geometric structure to resolve appearance ambiguities.}
%We also replaced FSAM with a standard self-attention layer, and the results are shown in `w/o FFT'.
%By operating in the frequency domain, FSAM bypasses the ambiguities of spatial textures and focuses on the underlying spectral signature of the glass surface, improving the model's performance. 
%Ablation on 3D priors reveals that using either depth or point cloud features in isolation fails to achieve optimal results. This underscores the geometric complementarity between these two modalities, which our GeGB effectively exploits.
%Finally, incorporating the edge loss $L_{edge}$ not only sharpens the segmentation boundaries but also guides the network to prioritize the structural integrity of the glass surface, yielding superior detection performance.

{\bf Training Formulations.} As reported in Tab.~\ref{tab:ablation_study_module} (middle part), we analyze different learning strategies. We first directly use the VGGT's original depth and point cloud predictions as supervision (denoted as ``w/o P\&D-C''). We also use our corrected depths and point clouds as supervision separately, denoted as ``w/o P-C'' and ``w/o D-C'', respectively. In addition, we also remove the glass surface supervision on the geometric features by removing $G^c$ from the $\mathcal{L}_{g}$ and $\mathcal{L}_{b}$ in Eq.~\ref{eq:loss} (denoted as ``w/o $G^c$''). Last, we remove the glass boundary delineation, which is denoted as ``w/o $L_{b}$''. These degraded results demonstrate the necessity of our training objective formulation. \lywre{Although the depth interpolation strategy brings marginal improvements, the consistent gains indicate that our model can effectively exploit more accurate 3D supervision when available.}

{\bf Geometry vs. Semantics.} While we demonstrate the superiority of geometric context over large-scale pre-trained diffusion priors, we are interested in whether it can outperform large-scale pre-trained semantic priors.
%very recently released
To this end, we replace our VGGT-based encoder with the SAM3 image encoder~\cite{carion2025sam3}, a large foundation model for \lywcr{concept} segmentation. Note that we also apply LoRA to the SAM3 encoder for a fair comparison (denoted as ``VGGT$\rightarrow$SAM3''). The results in Tab.~\ref{tab:ablation_study_module} (\tao{bottom} part) show that, despite being powerful for general object segmentation, the semantic contexts from SAM3 are still less effective than our geometric contexts for GSD.

%Removing the original attention layers of VGGT~\cite{wang2025vggt} degrades the encoder’s feature representation capability, leading to a noticeable performance drop (see `Baseline'). 
%Introducing LoRA~\cite{hu2022lora} substantially alleviates this issue and significantly improves overall performance (see `Baseline+LoRA'). 
%Since our model is compatible with feature tokens from various ViT-based encoders, 
%we conducted experiments using the original VGGT~\cite{wang2025vggt} backbone and the SAM3 image encoder~\cite{carion2025sam3}, both integrated with our baseline decoder. These configurations are respectively referred to as `VGGT+LoRA' and `SAM3+LoRA' in our ablation study. The additional attention layer in VGGT brings only a small improvement. This marginal gain suggests that the pre-trained backbone, enhanced by LoRA, already provides sufficiently rich semantic and geometric representations for the glass surface detection task. 
%Furthermore, ablating each module individually results in performance degradation, indicating that the proposed modules provide complementary benefits and jointly contribute to the final performance. 

\subsection{Runtime Efficiency}

We compare the number of parameters, GFLOPs, and the averaged inference time of our model, GhostingNet, GlassWizard, and our SAM3-based baseline, using a single NVIDIA RTX 4090 GPU. 
Tab.~\ref{tab:efficiency} reports the results, which show that compared with diffusion-based and SAM3-based methods, our method enjoys a reasonable runtime efficiency and can run interactively (around $8.5$ FPS).
\lywre{For details on the computational efficiency of each module, please refer to the supplementary materials.}
%Although more computationally intensive than compact models (GSDNet, RFENet, GhostingNet), our model remains highly efficient compared to SOTA methods: it reduces latency by 42ms relative to VGGT and is substantially faster than GlassWizard (177ms)~\cite{li2025glasswizard}.

%What's more, Further details of the efficiency of each module are available in the Appendix.

\begin{table}[ht]
% \small
\setlength{\tabcolsep}{6pt}
\centering
\caption{Runtime efficiency comparison on a single NVIDIA RTX 4090 GPU.}
\begin{tabular}{lccc}
\toprule
Methods & Param(M)&FLOPS(G)&Time(ms)\\
\midrule
%GSDNet&83.76&41.27&58.40\\
%RFENet&78.32&1,413.36&39.14\\
GhostingNet&271.53&321.70&32.94\\
GlassWizard&1189.73&890.78&177.00\\
SAM3-based&471.18& 2539.06&134.90\\
VGGT&989.17&1657.69&159.73\\
\rowcolor[gray]{.8}{\bf Ours}&684.37&1260.67&117.31\\

\bottomrule
\end{tabular}
\label{tab:efficiency}
\vspace{-3mm}
\end{table}

% {\flushleft\bf About depth correction.} 
% We investigate the impact of different ground-truth of depth maps and point clouds on network training and summarize the results in Tab.~\ref{tab:ablation_gt}. Here, `origin' denotes the unmodified depth maps and point clouds predicted by VGGT, while `fixed' refers to the depth maps and point clouds after applying the proposed depth correction algorithm. The results demonstrate that using consistent 3D information yields better performance; in contrast, inconsistent 3D information results in a noticeable performance drop. 

% \begin{table}[ht]
%   \centering
%   \setlength{\tabcolsep}{3pt}
%   % \footnotesize
%   \caption{Ablation study of the type of depth and point cloud maps.}
%   \begin{tabular}{ccccccc}
%     \toprule
%     Depth & Point & IoU↑ & F$_\beta$↑  & MAE↓ & BER↓ & ACC↑\\
%     \midrule
%     origin& origin& 93.04 & 0.965 &0.028 &2.36 &0.985 \\
%     fixed& origin& 92.84 & 0.963 &0.029 &2.43 &0.984 \\
%     origin& fixed& 92.76 & 0.962 &0.029 &2.40 &0.987 \\
%     \midrule
%     \multicolumn{2}{c}{Ours} &{\bf 93.25} & {\bf 0.968} & {\bf 0.025} & {\bf 2.18} & {\bf 0.989} \\
%     \bottomrule
%   \end{tabular}
%   \label{tab:ablation_gt}
%   \vspace{-3mm}
% \end{table}

% \subsection{Applications}

\begin{figure}[ht]
	\renewcommand{\tabcolsep}{0.8pt}
	\renewcommand\arraystretch{0.6}
    \renewcommand{\newsubwidth}{0.23}
        \centering
            \begin{tabular}{ccccc}
                % scene 1
                % \includegraphics[width=\newsubwidth\linewidth]{fig/exp/fail_case/GDD_5337/rgb.jpg}&
                % \includegraphics[width=\newsubwidth\linewidth]{fig/exp/fail_case/GDD_5337/depth.png}&
                % % \includegraphics[width=\newsubwidth\linewidth]{fig/exp/fail_case/GDD_5337/ghosting.png}&
                % \includegraphics[width=\newsubwidth\linewidth]{fig/exp/fail_case/GDD_5337/glasswizard.png}&
                % \includegraphics[width=\newsubwidth\linewidth]{fig/exp/fail_case/GDD_5337/ours.png}&
                % \includegraphics[width=\newsubwidth\linewidth]{fig/exp/fail_case/GDD_5337/gt.png}\\
  
                % scene 2
                \includegraphics[width=\newsubwidth\linewidth]{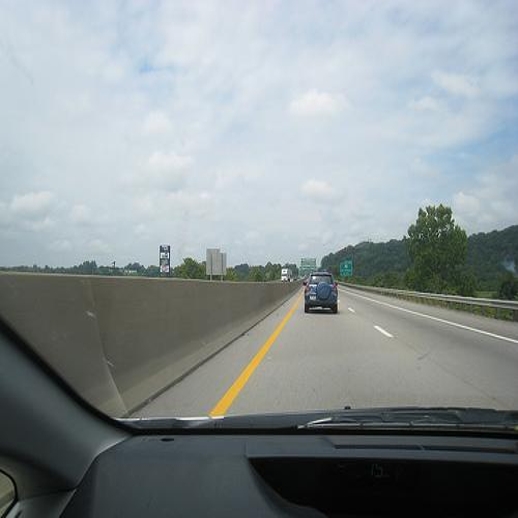}&
                \includegraphics[width=\newsubwidth\linewidth]{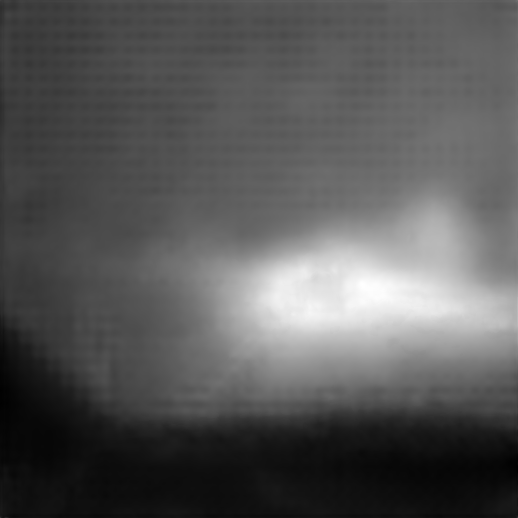}&
                \includegraphics[width=\newsubwidth\linewidth]{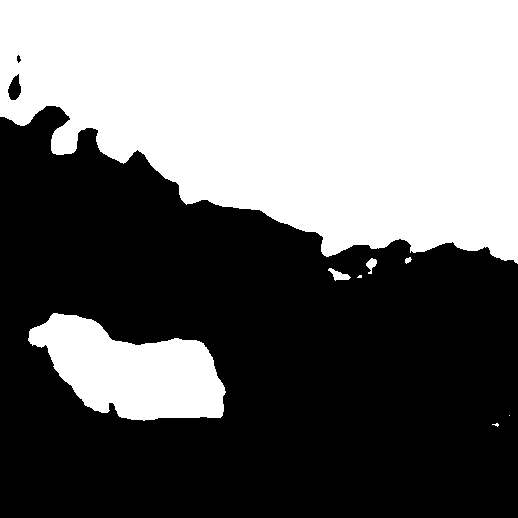}&
                \includegraphics[width=\newsubwidth\linewidth]{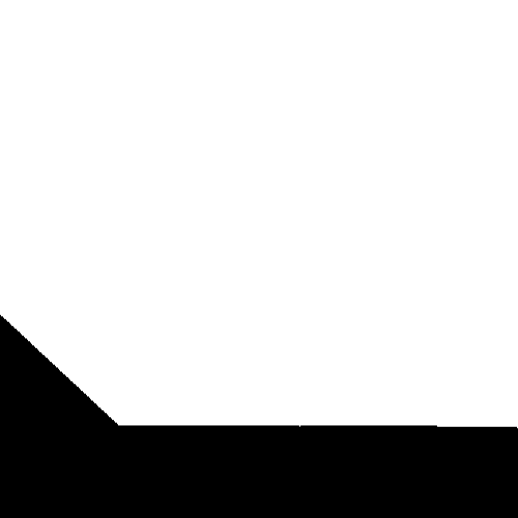}\\ 

                % annotation
                \small RGB&
                \small Depth&
                % \footnotesize Ghosting&
                %\footnotesize Glasswizard&
                \small Ours&
                \small GT
            \end{tabular}
    \vspace{-3mm}
    \caption{A failure case. Our method may fail when the glass surface (\eg, the windshield) is too close to the camera, providing insufficient surrounding geometric contexts.}
    \label{fig:failure_cases}
    \vspace{-3mm}
\end{figure}

\section{Conclusion}
In this paper, we have proposed a novel framework that grounds the glass surface detection task in 3D visual geometry. Our approach first exploits rich 3D prior knowledge from the visual geometry foundation model and generates pseudo-ground-truth depths and points in glass regions to strengthen their physical existence. We then formulate a multi-tasking training objective with a novel glass detection head. The glass detection head employs two core modules: FSAM for characterizing glass surfaces as high-frequency attenuation and GeGB for grounding 2D frequency features with 3D geometric features.
The proposed model has achieved new SOTA performance across seven standard benchmarks and demonstrated strong generalization capability. It can also run interactively (around $8.5$ FPS) on a single consumer-level GPU card.

%Our approach first applies a depth correction strategy and then performs end-to-end training to integrate 3D cues of glass surfaces into the detection process. In addition, we proposed two novel modules, FSAM and GeGB, to obtain unique glass features and fused features, respectively. Extensive experimental results demonstrate that our method achieves state-of-the-art performance across multiple benchmark datasets.

Our method does have limitations. While our model relies on high-quality 3D visual geometry cues, it may fail when the inferred geometric contexts are limited as shown in Fig.~\ref{fig:failure_cases}, where the windshield is too close to the camera.
%in the $1st$ scene, the background geometry closely aligns with the glass surface, leading the network to over-segment the entire region as glass. 
%since the glass area lacks significant features, it is more necessary to segment the windshield of the car through semantic understanding. However, our model incorporated incorrect geometric priors, resulting in a decrease in segmentation accuracy. Moreover, our model cannot fully handle situations such as an open glass door, which remains a subject for future investigation. Despite these limitations, our model still outperforms the competitors in overall performance.

% \section*{Impact Statement}
% This work advances the field of robust scene understanding by introducing a 3D geometry–grounded approach to glass surface detection. Potential positive applications include improved safety and reliability in robotic navigation and assistive technologies for the visually impaired. As with many perception systems, if deployed without proper testing, our method could contribute to errors in safety-critical environments. We encourage responsible validation and transparency in real-world deployment.

%This paper presents work whose goal is to advance the field of Machine Learning. There are many potential societal consequences of our work, none of which we feel must be specifically highlighted here.

\section*{Acknowledgments}
This work was supported by the National Natural Science Foundation of China (Grant No. 61902151). 
%%
%% The next two lines define the bibliography style to be used, and
%% the bibliography file.
\bibliographystyle{ACM-Reference-Format}
\balance
\bibliography{main}

\end{document}